\documentclass[10pt, a4paper]{scrartcl}

\usepackage{latexsym}
\usepackage{amssymb}
\usepackage[nonamelimits]{amsmath}
\usepackage[authoryear, round]{natbib}
\usepackage{url}
\usepackage[toc,page]{appendix}

\usepackage{color}
\definecolor{myred}{rgb}{0.5,0,0}
\definecolor{myblue}{rgb}{0,0,0.75}
\definecolor{mygreen}{rgb}{0,0.5,0}

\usepackage{ifpdf}

\ifpdf
   \usepackage[pdftex]{graphicx}
   \usepackage[pdftex,
               pdftitle={},
               pdfsubject={},
               pdfkeywords={},
               pdfauthor={Dirk Tasche},
               pdfstartview=FitR,
               breaklinks=true,
               colorlinks=true,
               citecolor=myred,
               linkcolor=myblue,
               urlcolor=mygreen]{hyperref}
\else
   \usepackage[dvips]{graphicx}
   \usepackage{hyperref}
   \usepackage{rotating}
\fi

\newtheorem{theorem}{Theorem}
\newtheorem{lemma}[theorem]{Lemma}

\newtheorem{proposition}[theorem]{Proposition}
\newtheorem{definition}[theorem]{Definition}

\newtheorem{assumption}[theorem]{Assumption}

\newcommand{\qed}{$\Box$}

\title{Fisher consistency for prior probability shift}

\author{%
Dirk Tasche\thanks{E-mail: dirk.tasche@gmx.net\newline
The author currently works at the Swiss Financial Market Supervisory Authority (FINMA).
The opinions expressed in this note are those of the author 
and do not necessarily reflect views of FINMA.}}

\date{}

\begin{document}

\maketitle

\begin{abstract}
We introduce Fisher consistency in the sense of unbiasedness as a 
desirable property for estimators of class prior probabilities. 
Lack of Fisher consistency could be used as a criterion
to dismiss estimators that are unlikely to deliver precise estimates
in test datasets under prior probability
and more general dataset shift. The usefulness of this unbiasedness concept
is demonstrated with three examples of classifiers used for quantification: Adjusted Classify \& Count, EM-algorithm
and CDE-Iterate.
We find that Adjusted Classify \& Count and EM-algorithm are Fisher consistent. A counter-example
shows that CDE-Iterate is not Fisher consistent and, therefore, cannot be trusted to
deliver reliable estimates of class probabilities.
\\[1ex]
\end{abstract}


\section{Introduction}
\label{se:intro}

The application of a classifier to a test dataset often is
based on the assumption that the data for training the classifier is representative of the test data. While this
assumption might be true sometimes or even most of the time, there may be circumstances when the distributions
of the classes or the features or both are genuinely different in the test set and the training set. Spam emails 
represent a familiar example of this situation: The typical contents of spam emails and their proportion of the
total daily number of emails received may significantly vary over time. Spam email filters developed on the
set of emails received last week may become less effective this week due to changes in the composition of the email traffic.
In the machine learning community, this phenomenon is called \textbf{dataset shift} or population drift. The area of research of
how to learn a model appropriate for a test set on a differently composed training set is called 
domain adaptation.

The simplest type of dataset shift occurs when the training set and the test set differ only in the distribution of
the classes of the instances. In this case, only the prior (or unconditional) class probabilities (or
\textbf{class prevalences}) change
between the training set and the test set; this type of dataset shift is called \textbf{prior probability shift}.
In the context of supervised or semi-supervised learning, the class prevalences in the labelled portion of the 
training set are always known.
In contrast, the class prevalences in the test set may be known or unknown at the time of the application of the classifier to the
test set, depending on the cause of the dataset shift. For example, in a binary classification exercise the
class prevalence of the majority class in the training set might deliberately be reduced by removing instances of that class at random,
in order to facilitate the training of the classifier. If the original training set were a random sample of the test set
then the test set class prevalences would be equal to the original training set class prevalences and therefore known. 
The earlier mentioned spam email filter problem represents an example of the situation where the test set class
prevalences are unknown due to possible dataset shift.

In this paper, we primarily study the question of how to estimate the unknown class prevalences in a test dataset
when the training set and the test set are related by prior probability shift. This problem was coined
\textbf{quantification} by \citet{forman2008quantifying}. Solutions to the problem are
known at least since the 1960s \citep{buck1966comparison} but research for more and better solutions has been ongoing ever since.
It seems, however, as if the criteria of what should be deemed a `good' solution are not fully clear.

In the words of \citet[][p.~287]{cox1974theoretical} \textbf{Fisher consistency} is described as ``Roughly this requires that if
the whole `population' of random variables is observed, then the method of estimation should give exactly the right answer''.
Obtaining the right answer when looking for the value of a parameter is an intuitive concept that exhibits 
similarity to the concept of unbiasedness. In particular,
if a wrong decision based on the estimated value of a parameter can entail severe financial loss or cause damage to a person's 
health, getting the value of the parameter right is very important. \citet{hofer2013drift} and \citet{tasche2014exact}  
present examples related to the estimation of credit losses. 
Nonetheless, requiring that the right 
answer be obtained also on finite samples from the population would be too harsh, given that by the nature of randomness 
the empirical distribution associated with a finite sample can be quite different from the population distribution.

For these reasons, we argue that as a minimum, it should be required that class prevalence
estimators (or \textbf{quantifiers}) be Fisher consistent. \citet[][p.~287]{cox1974theoretical} comment that
``Consistency, however, gives us no idea of the magnitude of the errors of estimation likely for any given $n$ 
[the sample size]''. Therefore, being Fisher consistent cannot be a sufficient criterion for a class prevalence
estimator to be useful. Rather, by the logic of a necessary criterion, lack of Fisher consistency should be considered a reason to dismiss 
a candidate estimator. For even for very large sample sizes, with such an estimator there would not be any guarantee 
of obtaining an approximation of the true parameter value.
 
Accordingly, focussing on \textbf{binary} classification and quantification, with this paper we make three contributions
to the literature on quantification of class prevalences:
\begin{itemize}
\item We formally introduce the concept of unbiasedness expressed as Fisher consistency, in a prior probability shift context and more generally for
quantification of class prevalences in the presence of any dataset shift.
\item We illustrate the usefulness of the notion of Fisher consistency, by demonstrating with three popular quantification approaches  
that Fisher consistency may be used as a criterion to cull inapt quantifiers. In other words, Fisher consistency can serve as 
a filter to eliminate quantifiers that are unlikely to provide precise estimates.
\item We show that Fisher consistency of an estimator is not a global concept that can be expected to hold
for all types of dataset shift.
To demonstrate this fact, we suggest a new type of dataset shift, called `invariant density ratio'-type
dataset shift, which generalises prior probability shift. 
We also propose a method for generating non-trivial examples  of this type of dataset shift.
\end{itemize}
`Invariant density ratio'-type dataset shift is interesting of its own for the following two reasons:
\begin{itemize}
\item It can be described as a covariate shift without the `contamination' effect that can entail the test set 
class prevalences to be very similar to the training set class prevalences \citep[][p.~83]{tasche2013art}.
\item Prior probability shift is a special case of `invariant density ratio'-type dataset shift (see 
Section~\ref{se:quant} below). Therefore, estimators which are Fisher consistent with respect to 
`Invariant density ratio'-type dataset shift potentially have a wider scope of application than 
estimators that are Fisher consistent only with respect to prior probability shift.
\end{itemize}
The plan for the paper is as follows:
\begin{itemize}
\item In Section~\ref{se:est}, we recall the concepts from binary classification and quantification that are
relevant for the subject of the paper. These concepts include Bayes classifiers and some types of dataset shift, including
prior probability shift. In addition, we motivate and define the notion of Fisher consistency.
\item In Section~\ref{se:approaches}, we describe three approaches to class prevalence estimation which serve to illustrate
the role of Fisher consistency for this task. The three approaches are Adjusted Classify \& Count \citep{forman2008quantifying},
the EM-algorithm \citep{saerens2002adjusting} and CDE-Iterate \citep{Xue:2009:QSC:1557019.1557117}.
\item In Section~\ref{se:Num}, we explore by numerical examples under which circumstances 
the three approaches discussed in Section~\ref{se:approaches} cease to be Fisher consistent for the estimation of class prevalences. The
most important finding here is that CDE-Iterate is not Fisher consistent under prior probability shift. Hence there is no 
guarantee that CDE-Iterate will find the true class prevalences even in the presence of that quite benign type of dataset shift. 
\item We conclude in Section~\ref{se:concl} with some comments on the findings of the paper.
\item Appendix~\ref{se:tables} presents some tables with computation results for additional information while Appendix~\ref{se:mixture} provides a 
mathematically rigorous derivation of the equation that characterises the limit of CDE-Iterate.
\end{itemize}
\textbf{Notation.} Concepts and notation used in this paper are formally introduced where needed. For additional quick reference,
here is a short list of the most important symbols:
\begin{itemize}
\item $\mathcal{X}$: Feature space.
\item $\{0, 1\}$: Two classes, 0 positive and 1 negative.
\item $(X, Y)$: $X$ is the vector of the features of an instance, $Y$ denotes the class of the instance.
\item $\mathrm{P}$: Probability distribution of the training set.
\item $\mathrm{Q}$: Probability distribution of the test set. 
\item $\mathrm{P}[Y=i\,|\,X]$: Feature-conditional probability of class $i$ in the training set (analogous for $\mathrm{Q}$).
\item $\mathrm{P}[X \in S\,|\, Y=i]$:  $\mathrm{P}[X\,|\, Y]$ denotes the class-conditional distribution of the features
in the training set. $\mathrm{P}[X \in S\,|\, Y=i]$ stands for the probability that the realisation of $X$ is an element of the
set $S$, conditional on the class of the instance being $i$ (analogous for $\mathrm{Q}$).
\item $g: \mathcal{X} \to \{0, 1\}$: Classifier that assigns class $g(x)$ to a realisation $x$ of the features of an instance.
\item $f_0, f_1$: Class-conditional feature densities on the training set.
\item $h_0, h_1$: Class-conditional feature densities on the test set.
\end{itemize}


\section{Estimating prior probabilities}
\label{se:est}

In this section, we recall some basic concepts from the theory of 
binary classification and quantification to build the basis for
the discussion of prevalence estimation approaches in Section~\ref{se:approaches} and
the numerical examples in Section~\ref{se:Num}. The concepts discussed
include Bayes classifiers and different types of dataset shift including
prior probability shift and its extension called here `invariant density ratio'-type
dataset shift.
In addition, we introduce
the notion of unbiasedness of prevalence estimators in the shape of Fisher consistency
which is most appropriate in the quantification context central for this paper.

\subsection{Binary classification and Bayes error at population level}
\label{se:Bayes}

The basic model we consider is a random vector $(X, Y)$ with values in a product $\mathcal{X} \times \{0,1\}$ of sets. 
For example, in applications for credit scoring, we might have $\mathcal{X} = \mathbb{R}^d$ for some positive integer $d$. 
A realisation of $X \in\mathcal{X}$ is interpreted as the vector of features of an observed instance.
$Y\in\{0,1\}$ is the class of the observed instance.
For the purpose of this paper, 0 is the interesting (positive) class, as in \citet{hernandez2012unified}.

\textbf{Classification problem.} On the basis of the observed features $X$ of an instance, make a guess (prediction) 
$g(X) \in \{0,1\}$ 
of the instance's class $Y$ such that the probability of an error is minimal. In mathematical terms: Find
$g^\ast: \mathcal{X} \to \{0,1\}$ such that
\begin{subequations}
\begin{equation}\label{eq:BayesErr}
\mathrm{P}[g^\ast(X) \not= Y]\ = \ \min\limits_{g} \mathrm{P}[g(X) \not= Y].
\end{equation}
The functions $g$ in \eqref{eq:BayesErr} used for predicting $Y$ are called (crisp) classifiers. The value of the minimum
in \eqref{eq:BayesErr} is called \textbf{Bayes error}.
\eqref{eq:BayesErr} accounts for two possibilities  to make classification errors: 
\begin{equation*}
 \begin{split}
 \text{`Predict 0 if the true class is 1'} & = \{g(X) = 0, Y=1\},\qquad\text{and}\\
 \text{`Predict 1 if the true class is 0'} & = \{g(X) = 1, Y=0\}.
 \end{split}
 \end{equation*} 
\textbf{Cost-sensitive errors.} In practice, the consequences of these two erroneous predictions might have different severities. 
Say the cost related to
`predict 0 if the true class is 1' is $c_1 \ge 0$, and the cost related to `predict 1 if the true class is 0' is $c_0 \ge 0$.
To deal with the issue of different severities, a cost-sensitive version of \eqref{eq:BayesErr} can be studied:
\begin{multline}\label{eq:costBayesErr}
c_1\,\mathrm{P}[g^\ast(X) = 0,\, Y=1] + c_0\,\mathrm{P}[g^\ast(X) = 1,\, Y=0] = \\ 
\min\limits_{g} c_1\,\mathrm{P}[g(X) = 0,\, Y=1] + c_0\,\mathrm{P}[g(X) = 1,\, Y=0].
\end{multline}
\end{subequations}
To make this problem non-trivial, of course one has to assume that $c_0 + c_1 > 0$. 

\textbf{Bayes classifier.} A solution $g^\ast$ to \eqref{eq:costBayesErr}
and therefore also to \eqref{eq:BayesErr} (case of $c_0 = c_1 = 1$)
exists and is well-known (see Section~2.2 of \citet{vanTrees} or Section~1.3 of \citet{Elkan01}):
\begin{equation}\label{eq:BayesClassifier}
g^\ast(X) \ = \ \begin{cases} 
    0, & \text{if}\ \mathrm{P}[Y=0\,|\,X] > \frac{c_0}{c_0+c_1},\\
    1, & \text{if}\ \mathrm{P}[Y=0\,|\,X] \le \frac{c_0}{c_0+c_1}.
    \end{cases}
\end{equation}
In this equation, $\mathrm{P}[Y=0\,|\,X]$ denotes the (non-elementary) conditional probability of the event $Y=0$
given $X$, as defined in standard textbooks on statistical learning and probability theory (see Appendix~A.7 of 
\citet{devroye1996probabilistic} and Section~4.1 of \citet{Durrett} respectively). Being a function 
of $X$, $\mathrm{P}[Y=0\,|\,X]$ is also a non-constant random variable whenever $X$ and $Y$ are not stochastically independent. 
In the following, we also call $\mathrm{P}[Y=0\,|\,X]$ \textbf{feature-conditional class probability}.
The function $g^\ast(X)$ as defined in \eqref{eq:BayesClassifier} is called a Bayes classifier.

A proof of
\eqref{eq:BayesClassifier} is also provided in Appendix~\ref{se:mixture} below (see Lemma~\ref{le:minimum}). That
proof shows, in particular, that the solution $g^\ast$ to \eqref{eq:costBayesErr} is unique in the sense of
$\mathrm{P}[g^\ast(X) = \tilde{g}(X)]=1$ for any other minimiser $\tilde{g}$ of \eqref{eq:costBayesErr}, as long as 
the distribution of the ratio of the class-conditional feature densities is continuous (see 
Section~\ref{se:quant} for the definition of the density ratio).

\subsection{Binary classification and Bayes error at sample level}
\label{se:BayesSample}

In theory, the binary classification problem with cost-sensitive errors is completely solved in Section~\ref{se:Bayes}.
In practice, however, there are issues that can make the Bayes classifier \eqref{eq:BayesClassifier} unfeasible:
\begin{itemize}
\item Typically, the joint probability distribution of $(X,Y)$ is not exactly known but has to be inferred from a finite
sample (called `training set') $(x_{1, tr}, y_{1, tr}), \ldots, (x_{m, tr}, y_{m, tr}) \in \mathcal{X} \times \{0,1\}$. 
If $\mathcal{X}$ is high-dimensional this may be difficult and require a large sample size to reduce errors due to
random variation.
\item As the Bayes classifier is explicitly given by \eqref{eq:BayesClassifier}, one can try to avoid estimating the
entire distribution of $(X,Y)$ and, instead, only estimate the feature-conditional probability $\mathrm{P}[Y=0\,|\,X]$ 
(also called posterior probability).
However, this task is not significantly simpler than estimating the distribution of $(X,Y)$ -- as illustrated by the
fact that methods for the estimation of non-elementary conditional probabilities constitute a major branch of 
applied statistics.
\end{itemize}
By structural assumptions on the nature of the classification problem \eqref{eq:costBayesErr}, the approach based on
direct estimation of the feature-conditional probability can be rendered more accessible. Logistic regression provides 
the possibly most important example for this approach \citep[see, e.g.,][]{Cramer2003}. But the price of the underlying
assumptions may be high and include significant deterioration of goodness of fit. That is why alternative approaches based on direct
implementation of the
optimisation of the right-hand side of  \eqref{eq:costBayesErr} are popular. \citet{lessmann2015benchmarking} 
give a survey of the variety of methods available, just for application to credit scoring.

As mentioned in the introduction, one of the topics of this paper is an investigation into the question of
whether certain quantifiers are Fisher consistent. In Section~\ref{se:Num} below, the demonstration of lack of Fisher 
consistency is based on counter-examples which are non-trivial but simple enough to allow for the
exact computation of the feature-conditional class probabilities and hence also the Bayes classifiers.

\subsection{Dataset shift}
\label{se:shift}

Even if one has succeeded in estimating a Bayes classifier or at least a reasonably good approximate Bayes classifier, 
issues may arise that spoil its effective deployment. Quite often, it is assumed that any instance with known features vector 
$x$ but unknown class $y$ that is presented for classification has been drawn at random from the same population as
the training set. However, for a number of reasons this assumption can be wrong 
\citep[see, e.g.,][]{quinonero2009dataset, kull2014patterns, MorenoTorres2012521, dal2015calibrating}.
A lot of research has been undertaken and is ongoing on methods to deal with this problem of so-called
dataset shift. In this paper, 
we consider the following variant of the problem:
\begin{itemize}
\item There is a training dataset $(x_{1, tr}, y_{1, tr}), \ldots, (x_{m, tr}, y_{m, tr}) \in \mathcal{X} \times \{0,1\}$
which is assumed to be an independent and identically distributed sample from the population distribution $\mathrm{P}(X,Y)$
of the random vector $(X,Y)$ as described in Section~\ref{se:Bayes}.
\item There is another dataset, the test dataset, 
$(x_{1, te}, y_{1, te}), \ldots, (x_{n, te}, y_{n, te}) \in \mathcal{X} \times \{0,1\}$
which is assumed to be an independent and identically distributed sample from a possibly different
population distribution $\mathrm{Q}(X,Y)$. 
Moreover, training and test data have been independently generated.
\item For the instances in the training dataset, their class labels are visible and can be made use of for learning
classifiers (i.e.\ solving optimisation problem \eqref{eq:costBayesErr}).
\item For the instances in the test dataset, their class labels are invisible to us or become visible only with large delay.
\end{itemize}
This assumption is intended to describe a setting where the test instances arrive batch-wise, not as a stream.
We also assume that the sizes $m$ of the training set and $n$ of the test set are reasonably large such that trying to infer 
properties of the respective 
population distributions makes sense. 

As far as the theory for this paper is concerned, we will ignore the issues caused by the fact that we know the training dataset 
distribution $\mathrm{P}(X,Y)$ and the test dataset distribution $\mathrm{Q}(X,Y)$ only by inference from finite samples.
Instead we will assume that we can directly deal with the population distributions $\mathrm{P}(X,Y)$ and $\mathrm{Q}(X,Y)$.
Throughout the whole paper, we make the assumption that there are both positive and negative instances in both of the
populations, i.e.\ it holds that
\begin{equation}
0 < \mathrm{P}[Y=0] < 1 \qquad \text{and}\qquad 0 < \mathrm{Q}[Y=0] < 1.
\end{equation}

The problem that $\mathrm{P}(X,Y)$ and $\mathrm{Q}(X,Y)$ may not be the same is treated under different names in the
literature \citep{MorenoTorres2012521}: dataset shift, domain adaptation, population drift and others.
There are several facets of the problem:
\begin{itemize}
 \item Classifiers may have to be adapted to the test set or re-trained.
 \item The feature-conditional probabilities may have to be adapted to the test set or re-estimated.
 \item The unconditional class probabilities (also called prior probabilities or prevalences) 
 may have to be re-adjusted for the test set or re-estimated.
\end{itemize} 
\textbf{Quantification.} In this paper, we focus on the estimation of the prevalences 
$\mathrm{Q}[Y=0]$ and $\mathrm{Q}[Y=1]$ in the test set, 
as parameters of distribution $\mathrm{Q}(X,Y)$. This problem is called
quantification \citep{forman2008quantifying} and of its own interest beyond its auxiliary function for classification and
estimation of the feature-conditional probabilities 
\citep{gonzalez2016quantification}.

\subsection{Quantification in the presence of prior probability shift}
\label{se:quant}

In technical terms, the quantification problem as presented in Section~\ref{se:shift} can be described as follows:
\begin{itemize}
\item We know the joint distribution of features and class labels $(X,Y)$ under the training set probability 
distribution $\mathrm{P}$.
\item We know the distribution of the features $X$ under the test set probability distribution $\mathrm{Q}$.
\item How can we infer the prevalences of the classes under $\mathrm{Q}$, 
i.e.\ the probabilities $\mathrm{Q}[Y=0]$ and $\mathrm{Q}[Y=1]= 1- \mathrm{Q}[Y=0]$, by making best possible use
of our knowledge of $\mathrm{P}(X,Y)$ and $\mathrm{Q}(X)$?
\end{itemize}
This question cannot be answered without assuming that the test set probability distribution $\mathrm{Q}$ `inherits' some
properties of the training set probability distribution  $\mathrm{P}$. This means to make more specific assumptions
about the structure of the dataset shift between training set and test set. In the literature, a variety of different
types of dataset shift have been discussed. See \citet{MorenoTorres2012521}, \citet{kull2014patterns} or 
\citet{hofer2015adapting} for a number
of examples, including `covariate shift' and `prior probability shift' which possibly are the two most studied types of
dataset shift. This paper focusses on prior probability shift.

\textbf{Prior probability shift.} This type of dataset shift also has been called `global shift' \citep{hofer2013drift}.
The assumption of prior probability shift is, in particular, appropriate for circumstances where the features of an instance
are caused by the instance's class membership \citep{fawcett2005response}.
Technically, prior probability shift can be described as 
`the class-conditional feature distributions of the training and test sets are the same', i.e.
\begin{equation}\label{eq:prior}
\mathrm{Q}[X\in S\,|\,Y=0] = \mathrm{P}[X\in S\,|\,Y=0] \quad \text{and}\quad
\mathrm{Q}[X\in S\,|\,Y=1] = \mathrm{P}[X\in S\,|\,Y=1],
\end{equation}
for all measurable sets $S \subset \mathcal{X}$. Note that \eqref{eq:prior} does \textbf{not} imply 
$\mathrm{Q}[X\in S] = \mathrm{P}[X\in S]$ for all measurable $S \subset \mathcal{X}$ because the training set 
class distribution $\mathrm{P}(Y)$ and the test set class distribution $\mathrm{Q}(Y)$ still can be different.

In this paper, we will revisit three approaches to the estimation of the prevalences in population $\mathrm{Q}$
in the presence of prior probability
shift as defined by \eqref{eq:prior}.  Specifically, we will check both in theory and by simple examples if these
three approaches satisfy the basic estimation quality criterion of Fisher consistency. For this purpose,
a slight generalisation of prior probability shift called `invariant density ratio'-type dataset shift will prove useful.
Before we introduce it, let us briefly recall some facts on conditional probabilities and probability densities.

\textbf{Feature-conditional class probabilities and class-conditional feature densities.} 
Typically, the class-conditional feature distributions $\mathrm{P}(X\,|\,Y=i)$,
$i = 0,1$, of a dataset have got densities $f_0$ and $f_1$ respectively with respect to some 
reference measure like the $d$-dimensional Lebesgue measure. Then also the unconditional feature distribution
$\mathrm{P}(X)$ has a density $f$ which can be represented as 
\begin{subequations}
\begin{equation}
  f(x) \ = \ \mathrm{P}[Y=0]\,f_0(x) + (1-\mathrm{P}[Y=0])\,f_1(x) \quad \text{for all}\ x \in \mathcal{X}.
\end{equation}
Moreover, the feature-conditional class probability  $\mathrm{P}[Y=0\,|\,X]$ can be expressed in terms of
the densities $f_0$ and $f_1$:
\begin{equation}\label{eq:probCond}
   \mathrm{P}[Y=0\,|\,X](x) \ = \ \frac{\mathrm{P}[Y=0]\,f_0(x)}
       {\mathrm{P}[Y=0]\,f_0(x) + (1-\mathrm{P}[Y=0])\,f_1(x)}, \quad x \in \mathcal{X}. 
\end{equation}
\end{subequations}
Conversely, assume that there is a density $f$ of the unconditional feature distribution $\mathrm{P}(X)$ and
the feature-conditional class probability  $\mathrm{P}[Y=0\,|\,X]$ is known. Then the class-conditional
feature densities $f_0$ and $f_1$ are determined as follows:
\begin{equation}\label{eq:derived}
\begin{split}
f_0(x) & = \frac{\mathrm{P}[Y=0\,|\,X](x)}{\mathrm{P}[Y=0]}\,f(x), \quad x \in \mathcal{X},\\
f_1(x) & = \frac{1-\mathrm{P}[Y=0\,|\,X](x)}{1-\mathrm{P}[Y=0]}\,f(x), \quad x \in \mathcal{X}.
\end{split}
\end{equation}

\textbf{Invariant density ratio.}  Assume that there are densities $f_0$ and $f_1$ for the
class-conditional feature distributions $\mathrm{P}(X\,|\,Y=i)$, $i = 0,1$, of the training set population.
We then say that the dataset shift from population $\mathrm{P}$ to $\mathrm{Q}$ is
of `invariant density ratio'-type if there are also densities $h_0$ and $h_1$ respectively for the 
class-conditional distributions $\mathrm{Q}(X\,|\,Y=i)$, $i = 0,1$, of the test set, and it holds that
\begin{equation}\label{eq:invariant}
\frac{f_0(x)}{f_1(x)}\ = \ \frac{h_0(x)}{h_1(x)}\quad \text{for all}\ x \in \mathcal{X}.
\end{equation}
Note that by \eqref{eq:derived}, the density ratio $\frac{f_0}{f_1}$ can be rewritten as
\begin{equation}\label{eq:ratio}
 \frac{f_0(x)}{f_1(x)} \ = \ \frac{\mathrm{P}[Y=0\,|\,X](x)}{1-\mathrm{P}[Y=0\,|\,X](x)}\,
                                 \frac{1-\mathrm{P}[Y=0]}{\mathrm{P}[Y=0]}, \quad x \in \mathcal{X},
\end{equation}
as long as $\mathrm{P}[Y=0\,|\,X](x) < 1$. Hence the density ratio can be calculated without knowledge of the values of the
densities if the feature-conditional class probabilities or reasonable approximations are known.

Obviously, `invariant density ratio' is implied by prior probability shift if all involved class-conditional
distributions have got densities. `Invariant density ratio'-type dataset shift was discussed in some detail
by \citet{tasche2014exact}. It is an interesting type of dataset shift for several reasons:
\begin{itemize}
\item[1)] `Invariant density ratio' extends the concept of prior probability shift.
\item[2)] Conceptually, `invariant density ratio' is similiar to covariate shift. 
To see this recall first that covariate shift is defined by the property that
the feature-conditional class probabilities of the training and test sets are the same
\citep[][Section~4.1]{MorenoTorres2012521}:
\begin{subequations}
\begin{equation}\label{eq:cov.shift}
\mathrm{P}[Y=0\,|\,X] \ =\ \mathrm{Q}[Y=0\,|\,X].
\end{equation}
Then, if there are densities of the class-conditional feature distributions under $\mathrm{P}$ and $\mathrm{Q}$
like for \eqref{eq:invariant}, \eqref{eq:cov.shift} can be rewritten as
\begin{equation}
\frac{\mathrm{P}[Y=0]}{1-\mathrm{P}[Y=0]}\,\frac{f_0(x)}{f_1(x)} \ =\ 
\frac{\mathrm{Q}[Y=0]}{1-\mathrm{Q}[Y=0]}\,\frac{h_0(x)}{h_1(x)}\quad \text{for all}\ x \in \mathcal{X}.
\end{equation}
\end{subequations}
Hence, covariate shift also can be described in terms of the ratio of the class-conditional densities, like 
`invariant density ratio'. The two types of dataset shift coincide if $\mathrm{P}[Y=0] = \mathrm{Q}[Y=0]$,
i.e.\ if the class prevalences in the training and test sets are the same. 
\item[3)] The maximum likelihood estimates of class prevalences under prior probability shift actually
are maximum likelihood estimates under `invariant density ratio' too 
\citep[][page 152 and Remark~1]{tasche2014exact}. Hence they can be calculated
with the EM (expectation-maximisation) algorithm as described by \citet{saerens2002adjusting}.
\item[4)] In terms of the dataset shift taxonomy of \citet{MorenoTorres2012521}, `invariant density ratio' 
is a non-trivial but manageable `other' shift, i.e.\ it is neither a prior-probability shift, nor a
covariate shift, nor a concept shift.
\end{itemize}
By properties 1) and 2), `invariant density ratio' has got conceptual similarities with both prior probability shift
and covariate shift. In Section~\ref{se:Num} below, 
we will make use of its manageability property 4) to construct an example of dataset shift that
reveals the limitations of some common approaches to the estimation of class prevalences.

\subsection{Fisher consistency}
\label{se:Fisher}

In Section~\ref{se:intro}, Fisher consistency of an estimator has been described as the property that 
the true value of a parameter related to a probability distribution is recovered when the estimator is applied 
to the distribution on the whole population. 
In this section, we apply the notion of Fisher consistency to the quantification of binary class prevalences. 

In practice, estimators are often Fisher consistent and asymptotically consistent (weakly or strongly, see
Section~10.4 of \citeauthor{VanDerVaart}, \citeyear{VanDerVaart}, for the definitions) at
the same time. \citet{gerow1989biometrics} discusses the questions of when this is the case and what concept
\citet{fisher1922mathematical} originally defined. Our focus on Fisher consistency is not meant to imply
that asymptotic consistency is a less important property. 
In the context of statistical learning, 
asymptotic consistency seems to have enjoyed quite a lot of attention, as shown for instance
by the existence of books like \citet{devroye1996probabilistic}. In addition, the convergence aspect
of asymptotic consistency often can be checked empirically by observing the behaviour of large samples.
However, in some cases it might be unclear if the limit of a seemingly asymptotically consistent 
large sample is actually the right one.
This is why, thanks to its similarity to the concept of unbiasedness which also refers 
to getting the right value of a parameter, Fisher consistency becomes an important
property. 

\begin{definition}[Fisher consistency]\label{de:Fisher}
In the dataset shift setting of this paper as described in Section~\ref{se:shift}, we say 
that an estimator $T(\mathrm{Q})$, applied to the elements $\mathrm{Q}$ of a 
family $\mathcal{Q}$ of possible population distributions 
of the test set, is Fisher consistent in $\mathcal{Q}$ for the prevalence of class 0 if it holds
that
\begin{equation}\label{eq:Fisher}
T(\mathrm{Q}) \ = \ \mathrm{Q}[Y=0] \quad \text{for all}\ \mathrm{Q} \in \mathcal{Q}.
 \end{equation} 
\end{definition}
This definition of Fisher consistency is more restrictive than the definition by \citet{cox1974theoretical}
quoted in Section~\ref{se:intro}.
For it requires the specification of a family of distributions to which the parameter `recovery' property applies.
The family $\mathcal{Q}$ of most interest for the purpose of this paper is the set of distributions $\mathrm{Q}$ that
are related to one fixed training set distribution $\mathrm{P}$ by prior probability shift, i.e.\ by \eqref{eq:prior}. 

\section{Three approaches to estimating class prevalences under prior probability shift}
\label{se:approaches}

In this section, we study three approaches to the estimation of binary class prevalences:
\begin{itemize}
\item Adjusted Classify \& Count (ACC) \citep{forman2008quantifying}, called `confusion matrix approach' by \citet{saerens2002adjusting},
but in use since long before \citep{buck1966comparison},
\item the EM (expectation maximisation) algorithm by \citet{saerens2002adjusting}, 
described before as maximum likelihood approach by \citet{peters1976numerical}, and
\item CDE-Iterate (CDE for class distribution estimation) by \citet{Xue:2009:QSC:1557019.1557117}.
\end{itemize}
A variety of other approaches have been and are being studied in the literature 
\citep[see the discussion in][for a recent overview]{hofer2015adapting}. The selection of approaches to be discussed 
in this paper was driven by findings of \citet{Xue:2009:QSC:1557019.1557117} and more recently \citet{karpov-porshnev-rudakov:2016:SemEval}.
According to that research, for the estimation of binary class prevalences the CDE-Iterate approach seems to
perform equally well or even stronger than ACC which by itself was found to outperform the popular EM-algorithm. 
In this section, we recall the technical details of the approaches which are 
needed to implement the numerical examples of Section~\ref{se:Num} below.

In addition, we check the three estimators on a theoretical basis for Fisher consistency. In particular with regard to 
CDE-Iterate, the theory is inconclusive with regard to its possible Fisher consistency. However, the example
of Section~\ref{se:2binormal} below shows that CDE-Iterate is not Fisher consistent for class 0 prevalence under
prior probability shift, in contrast to both ACC and the EM-algorithm.

\subsection{Adjusted Classify \& Count (ACC)}
\label{se:ACC}

Let $g: \mathcal{X} \to \{0,1\}$ be any classifier. Under prior probability shift as described by \eqref{eq:prior}, 
we then obtain
\begin{align}
\mathrm{Q}[g(X) = 0] & = \mathrm{Q}[Y=0]\,\mathrm{Q}[g(X)=0\,|\,Y=0] + (1-\mathrm{Q}[Y=0])\,\mathrm{Q}[g(X)=0\,|\,Y=1]
    \notag\\
    & = \mathrm{Q}[Y=0]\,\mathrm{P}[g(X)=0\,|\,Y=0] + (1-\mathrm{Q}[Y=0])\,\mathrm{P}[g(X)=0\,|\,Y=1].
    \label{eq:preACC}
\end{align}
If $\mathrm{P}[g(X)=0\,|\,Y=0] \not= \mathrm{P}[g(X)=0\,|\,Y=1]$, i.e.\ if $g(X)$ and $Y$ are not stochastically
independent, \eqref{eq:preACC} is equivalent to
\begin{equation}\label{eq:ACC}
\mathrm{Q}[Y=0] \ = \ \frac{\mathrm{Q}[g(X) = 0]-\mathrm{P}[g(X)=0\,|\,Y=1]}
    {\mathrm{P}[g(X)=0\,|\,Y=0]-\mathrm{P}[g(X)=0\,|\,Y=1]}.
\end{equation}
Equation~\eqref{eq:ACC} is called the ACC approach to the quantification 
of binary class prevalences.
Let us recall some useful facts about ACC:
\begin{itemize}
\item \eqref{eq:ACC} has been around for a long time, at least since the 1960s.
In this paper, the quantification approach related to \eqref{eq:ACC} is called `adjusted classify \& count' 
as in \citet{forman2008quantifying} because this term nicely describes what is done.
\item $\mathrm{Q}[g(X) = 0]$ is the proportion of instances in the test set (measured by counting)  
that are classified (predicted) positive by classifier $g(X)$. 
\item $\mathrm{P}[g(X)=0\,|\,Y=1]$ is the `false positive rate', as measured  for classifier
$g(X)$ on the training set.
\item $\mathrm{P}[g(X)=0\,|\,Y=0]$ is the `true positive rate', as measured for classifier
$g(X)$ on the training set.
\item \citet{forman2008quantifying} discusses ACC in detail and provides a number of variations of the theme 
in order to account for its deficiencies.
\item Possibly, the main issue with ACC is that in practice the result of the right-hand side of \eqref{eq:ACC}
can turn out to be negative or greater than 1. This can happen for one or both of the following two reasons:
\begin{enumerate}
\item The dataset shift in question actually is no prior probability shift, i.e.\ \eqref{eq:prior} does not hold.
\item The estimates of the true positive and true negative rates are inaccurate.
\item The estimation of the scoring function underlying the classifier from limited
training data may be inaccurate (both biased and subject to high variance).
\end{enumerate}
\item In theory, if the dataset shift is indeed a prior probability shift, the result of the right-hand side
of \eqref{eq:ACC} should be the same, regardless of which admissible (i.e.\ such that the denominator is not 
zero) classifier is deployed for determining the proportion of instances in the test set classified positive. Hence,
whenever in practice different classifiers give significantly different results, that could suggest that the assumption
of prior probability shift is wrong.
\item As long as $g(X)$ and $Y$ are at least somewhat dependent, possible lack of power of the classifier $g$ should not be 
an issue for the applicability of \eqref{eq:ACC} because the denominator on the right-hand side of \eqref{eq:ACC} is then
different from zero.
\end{itemize}
For a training set distribution $\mathrm{P}$ denote by $\mathcal{Q}_{\textrm{prior}} = \mathcal{Q}_{\textrm{prior}}(\mathrm{P})$
the family of distributions $\mathrm{Q}$ that are related to $\mathrm{P}$ by prior probability shift
in the sense of Definition~\ref{de:Fisher}, i.e.
\begin{equation}\label{eq:priorFamily}
    \mathcal{Q}_{\textrm{prior}} \ = \
\{\mathrm{Q}: \mathrm{Q}\ \text{is probability measure satisfying \eqref{eq:prior}}\}.
\end{equation}
Then, for fixed training set distribution $\mathrm{P}$ and fixed classifier $g(X)$
such that $g(X)$ and $Y$ are not independent under $\mathrm{P}$, the ACC approach is Fisher consistent 
in $\mathcal{Q}_{\textrm{prior}}$ for the prevalence of class 0 by construction:
Define the operator $T = T_{g,\mathrm{P}}$ by 
\begin{equation*}
T(\mathrm{Q}) \ = \ \frac{\mathrm{Q}[g(X) = 0]-\mathrm{P}[g(X)=0\,|\,Y=1]}
    {\mathrm{P}[g(X)=0\,|\,Y=0]-\mathrm{P}[g(X)=0\,|\,Y=1]}.
\end{equation*}
Then \eqref{eq:ACC} implies \eqref{eq:Fisher} for $\mathrm{Q} \in \mathcal{Q}_{\textrm{prior}}$. 
However, denote -- again for some fixed
training set distribution $\mathrm{P}$ -- by $\mathcal{Q}_{\textrm{invariant}} = \mathcal{Q}_{\textrm{invariant}}(\mathrm{P})$
the family of distributions $\mathrm{Q}$ that are related to $\mathrm{P}$ by `invariant 
density ratio'-type dataset shift
in the sense of \eqref{eq:invariant}, i.e.
\begin{equation}\label{eq:invariantFamily}
    \mathcal{Q}_{\textrm{invariant}} \ = \
\{\mathrm{Q}: \mathrm{Q}\ \text{is probability measure satisfying \eqref{eq:invariant}}\}.
\end{equation} 
Then in general the ACC approach is not Fisher consistent in $\mathcal{Q}_{\textrm{invariant}}$
as it is shown in Section~\ref{se:binormalInvariant} below that there are a classifier $g^\ast$, 
a distribution $\mathrm{P}^\ast$ and
a related distribution $\mathrm{Q}^\ast \in
\mathcal{Q}_{\textrm{invariant}}(\mathrm{P}^\ast)$ such that 
$$T_{g^\ast, \mathrm{P}^\ast}(\mathrm{Q}^\ast) \not= \mathrm{Q}^\ast[Y=0].$$

\subsection{EM-algorithm}
\label{se:EM}

\citet{saerens2002adjusting} made the EM-algorithm popular for the estimation of class prevalences, as a necessary
step for the re-adjustment of thresholds of soft classifiers. A closer inspection of the article by \citet{peters1976numerical}
shows that they actually had studied the same algorithm and provided conditions for its convergence. In particular,
this observation again draws attention to the fact that the EM-algorithm, deployed on data samples, 
should result in unique  maximum likelihood estimates of the class prevalences. 

As noticed by \citet{duPlessis2014110},
the population level equivalent of sample level maximum likelihood estimation of the class prevalences under an assumption of
prior probability shift is 
minimisation of the Kullback-Leibler distance between the estimated test set feature distribution and the
observed test set feature distribution. Moreover, \citet{tasche2014exact} observed that
the EM-algorithm finds the true values of the class prevalences not only under prior probability
shift but also under `invariant density ratio'-type dataset shift. In other words, the EM-algorithm
is Fisher consistent both in $\mathcal{Q}_{\textrm{prior}}$ and $\mathcal{Q}_{\textrm{invariant}}$, as 
defined in \eqref{eq:priorFamily} and \eqref{eq:invariantFamily} respectively, for the prevalence of class 0.
See Proposition~\ref{pr:invariant} below for a formal proof.

In the case of two classes, the maximum-likelihood version of the EM-algorithm in the sense of solving the likelihood equation
is more efficient than the EM-algorithm itself. This statement applies even more to the population level calculations.
In this paper, therefore, we describe the result of the EM-algorithm as the unique solution of a specific equation, 
as described by \citet{tasche2014exact}. 

\textbf{Calculating the result of the EM-algorithm.} 
In the population setting of Section~\ref{se:shift} with training set distribution $\mathrm{P}(X,Y)$ and 
test set distribution $\mathrm{Q}(X,Y)$, assume that the class-conditional feature distributions $\mathrm{P}(X\,|\,Y=i)$,
$i = 0,1$, of the training set have got densities $f_0$ and $f_1$. Define the density ratio $R$ by
\begin{equation}\label{eq:densRatio}
R(x) \ = \ \frac{f_0(x)}{f_1(x)}, \quad \text{for}\ x \in \mathcal{X}.
\end{equation}
We then define the estimation operator $T_R(\mathrm{Q})$ for the prevalence of class 0 as the unique solution $q \in
(0,1)$ of the equation \citep{tasche2014exact}
\begin{subequations}
\begin{equation}\label{eq:EM.result}
0 \ = \ \mathrm{E}_{\mathrm{Q}}\left[\frac{R(X)-1}{1 + q\,(R(X)-1)}\right], 
\end{equation}
where $\mathrm{E}_{\mathrm{Q}}$ denotes the expectation operator with respect to $\mathrm{Q}$. Unfortunately,
not always does a solution of \eqref{eq:EM.result} exist in $(0,1)$. There exists a solution in $(0,1)$
if and only if 
\begin{equation}
\mathrm{E}_{\mathrm{Q}}[R(X)] > 1 \quad \text{and}\quad \mathrm{E}_{\mathrm{Q}}\left[\frac1{R(X)}\right] > 1,
\end{equation}
and if there is a solution in $(0,1)$ it is unique \citep[][Remark~2(a)]{tasche2014exact}.
\end{subequations}
\begin{proposition}\label{pr:invariant}
The operator $T_R(\mathrm{Q})$ (EM-algorithm), as defined by \eqref{eq:EM.result}, is Fisher consistent in 
$\mathcal{Q}_{\textrm{prior}}$ and $\mathcal{Q}_{\textrm{invariant}}$ for the prevalence of class 0.
\end{proposition}
\textbf{Proof.} We only have to prove the claim for $\mathcal{Q}_{\textrm{invariant}}$ because $\mathcal{Q}_{\textrm{prior}}$ is
a subset of $\mathcal{Q}_{\textrm{invariant}}$. Let any $\mathrm{Q} \in \mathcal{Q}_{\textrm{invariant}}$ be given and
denote by $h_i$, $i=0,1$, its class-conditional feature densities.
By \eqref{eq:invariant} and \eqref{eq:probCond}, it then follows that
\begin{align*}
\mathrm{E}_{\mathrm{Q}}\left[\frac{R(X)-1}{1 + \mathrm{Q}[Y=0]\,(R(X)-1)}\right] & =
      \mathrm{E}_{\mathrm{Q}}\left[\frac{h_0-h_1}{h_1 + \mathrm{Q}[Y=0]\,(h_0-h_1)}\right] \\
  & = \frac{\mathrm{E}_{\mathrm{Q}}\bigl[\mathrm{Q}[Y=0\,|\,X]\bigr]}{\mathrm{Q}[Y=0]} -
  \frac{\mathrm{E}_{\mathrm{Q}}\bigl[\mathrm{Q}[Y=1\,|\,X]\bigr]}{\mathrm{Q}[Y=1]}\\
  & = 1 - 1 = 0.
\end{align*}
Hence, the prevalence $\mathrm{Q}[Y=0]$ of class 0 is a solution of \eqref{eq:EM.result} and, therefore, the only
solution. As a consequence, $T_R(\mathrm{Q})$ is well-defined and satisfies $T_R(\mathrm{Q}) = \mathrm{Q}[Y=0]$. 
\hfill \qed

In practice, for fixed $q$ the right-hand side of \eqref{eq:EM.result} could be estimated on a test set sample 
$(x_{1, te}, y_{1, te}), \ldots$, $(x_{n, te}, y_{n, te})$ as in Section~\ref{se:shift} by the sample
average
$$
\frac{1}{n} \sum_{i=1}^n \frac{R(x_{i, te})-1}{1 + q\,(R(x_{i, te})-1)},
$$
where $R$ could be plugged in as a training set estimate of the density ratio by means of \eqref{eq:ratio}.

\subsection{CDE-Iterate}
\label{se:CDE-Iterate}

In order to successfully apply the ACC quantification approach as described in Section~\ref{se:ACC}, we must 
get hold of reliable estimates of the training set true and false positive rates of the classifier deployed. 
If the positive class
is the minority class, the estimation of the true positive rate can be subject to large uncertainties and, hence,
may be hard to achieve with satisfactory accuracy. Similarly, if the negative class is the minority class, the
estimation of the false positive rate can be rather difficult. 

Application of the EM-algorithm as
introduced by \citet{saerens2002adjusting} or described in Section~\ref{se:EM} requires reliable estimation
of the feature-conditional class probabilities or the density ratio. Again, such estimates in general are hard to achieve.
That is why alternative methods for quantification are always welcome. In particular, methods that are based exclusively 
on learning one or more crisp classifiers are promising. For learning classifiers is a well-investigated problem
for which efficient solution approaches are available \citep[see, for instance,][for a survey related to
credit scoring]{lessmann2015benchmarking}.

\citet{Xue:2009:QSC:1557019.1557117} proposed `CDE-Iterate' (CDE for class distribution estimation) which is 
appropriately summarised by
\citet{karpov-porshnev-rudakov:2016:SemEval} as follows: ``The main idea of this method is to retrain a classifier
at each iteration, where the iterations progressively improve
the quantification accuracy of performing the
`classify and count' method via the generated cost-sensitive classifiers.'' \citeauthor{Xue:2009:QSC:1557019.1557117} 
motivated the CDE-Iterate algorithm as kind of an equivalent of the EM-algorithm, with the training
set feature-conditional class probabilities replaced by Bayes classifiers (or approximation of the classifiers) learnt
on the training set.

In this paper, we do not retrain 
a classifier but make use of the fact that we have got a closed-form representation of the 
optimal classifier resulting from the retraining, by \eqref{eq:BayesClassifier}. 
Taking this into account and using notation adapted for this paper, we obtain the following 
description of the CDE-Iterate procedure:
\begin{itemize}
\item[] \textbf{CDE-Iterate algorithm}
\item[1)] Set initial parameters: $k = 0$, $c_0^{(0)}=1$, $c_1^{(0)}=1$.
\item[2)] Find Bayes classifier under training distribution $\mathrm{P}(X,Y)$: 
$$g_k(X) \ = \ \begin{cases} 
    0, & \text{if}\ \mathrm{P}[Y=0\,|\,X] > \frac{c_0^{(k)}}{c_0^{(k)}+c_1^{(k)}},\\
    1, & \text{if}\ \mathrm{P}[Y=0\,|\,X] \le \frac{c_0^{(k)}}{c_0^{(k)}+c_1^{(k)}}.
    \end{cases}$$
\item[3)] Under test feature distribution $\mathrm{Q}(X)$ compute
$q_k  =  \mathrm{Q}[g_k(X) = 0]$.
\item[4)] Increment $k$ by 1.
\item[5)] Reset cost parameters: 
$c_1^{(k)} =  \frac{1-q_k}{1-\mathrm{P}[Y=0]}$, $c_0^{(k)} =  \frac{q_k}{\mathrm{P}[Y=0]}$.
\item[6)] If convergence is reached or $k = k_{\max}$ then stop, and accept $q_k$ as 
the CDE-Iterate estimate of $\mathrm{Q}[Y=0]$. Else continue with step 2.
\end{itemize}
\citet{Xue:2009:QSC:1557019.1557117} did not provide a proof of convergence or unbiasedness for CDE-Iterate.
In Section~6 of their paper, they state that
``one improvement would be to adapt the CDE-Iterative method to automatically terminate once the class distribution estimate converges. 
This might improve overall performance over any specific CDE-Iterate-$n$ method and would eliminate 
the problem of identifying the appropriate number of iterations. It is possible that such a CDE-converge method would outperform CDE-AC.'' 
In this paper, we prove convergence of CDE-Iterate and also show
that it does not outperform `CDE-AC' (Adjusted Classify \& Count in the notation of this paper) 
for class distribution estimation under prior probability shift, thus answering
the quoted research questions of \citeauthor{Xue:2009:QSC:1557019.1557117}.

The proof of the convergence of CDE-Iterate as described above is provided in Proposition~\ref{pr:CDE} in Appendix~\ref{se:mixture} below.
There it is also shown that the limit $q^\ast = \lim_{k\to\infty} q_k$
solves the following equation\footnote{%
Subject to the technical condition that $\mathrm{P}(X)$ has a density $f$ such
that $\mathrm{Q}[f(X)>0] =1$.}:
\begin{equation}\label{eq:limit}
q^\ast\ =\ \begin{cases}
\mathrm{Q}\left[R(X) \ge \frac{1-q^\ast}{q^\ast}\right], & \text{if}\ q_0 \ge q_1 \ \text{and}\ q_k > q^\ast\ 
        \text{for all}\ k,\\[1ex]
\mathrm{Q}\left[R(X) > \frac{1-q^\ast}{q^\ast}\right], & \text{otherwise}, 
\end{cases}      
\end{equation}
where $R(x)$ is defined by \eqref{eq:densRatio}.

The limit result \eqref{eq:limit} is quite general in so far as it is not based on any assumption with 
regard to the type of dataset shift between the training set distribution $\mathrm{P}(X,Y)$ and the test set
distribution $\mathrm{Q}(X,Y)$. If we restrict the type of dataset shift to prior probability shift or
`invariant density ratio'-type dataset shift as defined in Section~\ref{se:quant}, does \eqref{eq:limit} then
imply Fisher consistency in either of these two families of distributions for the prevalence of class 0?
As we show by example in Section~\ref{se:Num} following, the answer to this question is `no'.

\section{Numerical examples}
\label{se:Num}

In Section~\ref{se:approaches}, we have found that ACC as an estimator of class prevalences is Fisher consistent 
for prior probability shift while the EM-algorithm is even Fisher consistent for the more general `invariant density ratio'-type 
dataset shift. We have not yet answered the question if CDE-Iterate is Fisher consistent for either of these two 
dataset shift types. 

The property of an estimator to be Fisher consistent is something that has to be proved. In contrast, lack of Fisher
consistency of an estimator is conveniently shown by providing a counter-example. This is the purpose of the following subsections: We
show by examples that
\begin{itemize}
\item CDE-Iterate is not Fisher consistent for prior probability shift (and hence for `invariant density ratio'-type dataset shift
neither),
\item ACC is not Fisher consistent for `invariant density ratio'-type dataset shift, and
\item the EM-algorithm is no longer Fisher consistent if the `invariant density ratio'-type dataset shift is slightly modified.
\end{itemize}

We present the counter-examples as a simulation and estimation experiment 
that is executed for each of the three following example models:
\begin{itemize}
\item Section \ref{se:2binormal}: Binormal model with equal variances for both training and test set (prior probability shift).
\item Section \ref{se:binormalInvariant}: Binormal model with equal variances for training set and model with non-normal
class-conditional densities but identical density ratio
for test set (`invariant density ratio'-type dataset shift).
\item Section \ref{se:squareRoot}: Binormal model with equal variances for training set and model with non-normal
class-conditional densities and a different density ratio for test set (neither prior probability shift nor 
`invariant density ratio'-type dataset shift).
\end{itemize}
The experimental design is described in Section~\ref{se:experiment} below. The classical binormal model with equal variances 
has been selected as the training set model for the following reasons:
\begin{itemize}
\item Logistic regression finds the correct feature-conditional class probabilities.
\item Needed algorithms are available in common software packages like R.
\item The ratio of the class-conditional feature densities has a particularly simple shape, see \eqref{eq:ratioBinorm} below.
\item The binormal model with equal variances has been found useful before for a similar experiment
\citep{tasche2016quantification}.
\end{itemize}
\textbf{Thresholds for Bayes classifier under dataset shift.} 
We deploy the logistic regression as coded by \citet{RSoftware}. Therefore, it is convenient to always
use the feature-conditional class-probability $\mathrm{P}[Y=0\,|\,X]$ as the Bayes classifier, both for
the training set and for the test set after a possible dataset shift (then with a modified threshold).
In order to be able to do so, we observe that, under prior probability shift or even `invariant density ratio'-type dataset shift, 
the test set Bayes classifier $g_{\mathrm{test}}(X)$ for the cost-sensitive error 
criterion \eqref{eq:costBayesErr} can be represented both as
\begin{align}
g_{\mathrm{test}}(X) & \ \stackrel{\text{\eqref{eq:BayesClassifier}}}{=} \ \begin{cases} 
    0, & \text{if}\ \mathrm{Q}[Y=0\,|\,X] > \frac{c_0}{c_0+c_1},\\
    1, & \text{if}\ \mathrm{Q}[Y=0\,|\,X] \le \frac{c_0}{c_0+c_1},
    \end{cases}\notag\\
\intertext{and as \citep[see][Section 2.2]{saerens2002adjusting}}
g_{\mathrm{test}}(X) & \ = \ \begin{cases} 
    0, & \displaystyle{}\text{if}\ \mathrm{P}[Y=0\,|\,X] > \frac{c_0\,\frac{1-\mathrm{Q}[Y=0]}{1-\mathrm{P}[Y=0]}}
        {c_0\,\frac{1-\mathrm{Q}[Y=0]}{1-\mathrm{P}[Y=0]} +c_1\,\frac{\mathrm{Q}[Y=0]}{\mathrm{P}[Y=0]}},\\
    \ & \ \\
    1, & \displaystyle{}\text{if}\ \mathrm{P}[Y=0\,|\,X] \le \frac{c_0\,\frac{1-\mathrm{Q}[Y=0]}{1-\mathrm{P}[Y=0]}}
        {c_0\,\frac{1-\mathrm{Q}[Y=0]}{1-\mathrm{P}[Y=0]} +c_1\,\frac{\mathrm{Q}[Y=0]}{\mathrm{P}[Y=0]}}.
    \end{cases} \label{eq:transform}
\end{align}

\subsection{Design of the experiment}
\label{se:experiment}

In the subsequent sections \ref{se:2binormal}, \ref{se:binormalInvariant} and \ref{se:squareRoot}, we conduct
the following experiment and report its results\footnote{%
The R-scripts used for creating the tables and figures of this paper can be received upon request
from the author.}:
\begin{itemize}
\item For a given training set (sample and population) represented by distribution $\mathrm{P}(X,Y)$, 
we determine the Bayes classifier that is optimal for minimising the 
Bayes error \eqref{eq:BayesErr}. We represent the Bayes classifier by a decision threshold applied to the feature-conditional
probability of class 0 as in \eqref{eq:BayesClassifier}, with $c_0 = 1 = c_1$.
\item We create test sets (by simulation or as population distribution), represented by distributions $\mathrm{Q}(X,Y)$, which
are related to the training set by certain types of dataset shift, including prior probability shift and `invariant density ratio'-type
dataset shift. 
\item On the test sets, we deploy three different quantification methods for the estimation of the prevalence of class 0 
(see also Definition~\ref{de:acronyms} below): 
    CDE-Iterate (defined in Section~\ref{se:CDE-Iterate}), Adjusted Classify \& Count (defined in Section~\ref{se:ACC}),
    and EM-algorithm (defined in Section~\ref{se:EM}).
\item Based on the estimated class 0 prevalences, we adapt the threshold of the Bayes classifier according to 
\eqref{eq:transform} such that it would
be optimal for minimising the Bayes error (equivalently for maximising the classification accuracy) if the estimated prevalence were
equal to the true test set class 0 prevalence and the test set were related to the training set by prior probability shift.
\item We report the following results for samples and populations:
\begin{itemize}
\item Classification accuracy and F-measure (see \eqref{eq:accuracy} and \eqref{eq:Fmeasure} for the definitions) 
of the adapted Bayes classifier when applied to the test set, 
because these measures were used by \citet{Xue:2009:QSC:1557019.1557117}.
\item Estimated prevalences of class 0, for direct comparison of estimation results and true values.
\item Relative error: If $q$ is the true probability and $\tilde{q}$ the estimated probability, then
we tabulate
\begin{equation}\label{eq:relError}
    \max\left(\frac{|\tilde{q}-q|}{q}, \frac{|1-\tilde{q}-(1-q)|}{1-q}\right) \ = \ \frac{|\tilde{q}-q|}{\min(q, 1-q)}.
\end{equation}
Relative error behaves similar to Kullback-Leibler distance used by \citet{karpov-porshnev-rudakov:2016:SemEval}, but is defined also 
for $\tilde{q} = 0$ and $\tilde{q}=1$ and, moreover, has a more intuitive interpretation.
\end{itemize}
\end{itemize}
\textbf{Modelled class prevalences.} The setting is broadly the same as for the artificial dataset in \citet{karpov-porshnev-rudakov:2016:SemEval}. 
For each model, we consider a training set with class probabilities 50\%, combined with
test sets with class 0 probabilities 1\%, 5\%, 10\%, 30\%, 50\%, 70\%, 90\%, 95\% and 99\%.
For the samples  as well as for the population distributions, 
we deploy the estimation approaches whose acronyms are given in the following definition to estimate the test set class 0 prevalences.
\begin{definition}[Acronyms for estimation approaches]\label{de:acronyms}\ \\
The following estimation approaches are used in this section:
\begin{itemize}
\item CDE-Iterate in three variants:
\begin{itemize}
\item CDE1: First iteration of the algorithm described in Section~\ref{se:CDE-Iterate}. Identical with
Classify \& Count of \citet{forman2008quantifying}.
\item CDE2: Second iteration of the algorithm described in Section~\ref{se:CDE-Iterate}.
\item CDE$\infty$: CDE-Iterate converged, as described in Section~\ref{se:CDE-Iterate}.
\end{itemize}
\item ACC: Adjusted Classify \& Count as described in Section~\ref{se:ACC}.
\item EM: EM-algorithm as described in Section~\ref{se:EM}.
\end{itemize}
\end{definition}

\subsection{Training set: binormal; test set: binormal}
\label{se:2binormal}


We consider the classical binormal model with equal variances as 
an example that fits well into the prior probability shift setting of Section~\ref{se:quant}. We specify the binormal model by
defining the class-conditional feature distributions.
\begin{subequations}
\begin{itemize}
\item \textbf{Training set:} Both class-conditional feature distributions are normal, with equal variances, i.e.
\begin{equation}\label{eq:CondNormal}
    \mathrm{P}(X\,|\,Y=0)  = \mathcal{N}(\mu, \sigma^2),\qquad
    \mathrm{P}(X\,|\,Y=1)  = \mathcal{N}(\nu, \sigma^2), 
\end{equation}
with $\mu < \nu$ and $\sigma > 0$.
\item \textbf{Test set:} Same as training set.
\item For this section's numerical experiment, the following parameter values have been chosen:
\begin{equation}\label{eq:values}
\mu = 0, \qquad \nu = 2, \qquad \sigma = 1.
\end{equation} 
\end{itemize}
\end{subequations}
For the sake of brevity, in the following we sometimes refer to the setting of this 
section as `double' binormal. The feature-conditional class probability $\mathrm{P}[Y=0\,|\,X]$ in the training set is given by
\begin{subequations}
\begin{equation}\label{eq:testCondProb}
    \mathrm{P}[Y=0\,|\,X](x) \  = \ \frac{1}{1 + \exp(a\,x + b)}, \quad x \in \mathbb{R},
\end{equation}
with $a = \frac{\nu-\mu}{\sigma^2} > 0$ and $b = \frac{\mu^2-\nu^2}{2\,\sigma^2} + 
\log\left(\frac{1-\mathrm{P}[Y=0]}{\mathrm{P}[Y=0]}\right)$. For the density ratio $R$ according
to \eqref{eq:densRatio}, we obtain
\begin{equation}\label{eq:ratioBinorm}
R(x) \ = \ \exp\left(x\,\tfrac{\mu-\nu}{\sigma^2} + \tfrac{\nu^2-\mu^2}{2\,\sigma^2}\right),
\quad x \in \mathbb{R}.
\end{equation}
\end{subequations}

For the sample version of the example in this section, we create by Monte-Carlo simulation a training sample
$\bigl((x_{1, tr}, y_{1, tr}), \ldots, 
    (x_{m, tr}, y_{m, tr})\bigr)
\in (\mathbb{R}\times\{0,1\})^m$ and test samples $\bigl((x_{1, te}, y_{1, te}),$ $\ldots,$ $(x_{n, te}, y_{m, te})\bigr)
\in (\mathbb{R}\times\{0,1\})^n$
with class-conditional feature distributions given by \eqref{eq:CondNormal} that
approximate the training set population distribution $\mathrm{P}$ and the test set population distributions 
$\mathrm{Q}$ as described in general terms in Section \ref{se:quant} and more specifically here by \eqref{eq:CondNormal},
\eqref{eq:values} and
the respective class 0 prevalences.
\begin{itemize}
\item In principle, $(x_{1, tr}, y_{1, tr}), \ldots, (x_{m, tr}, y_{m, tr})$ is an
independent and identically distributed (iid) sample from $\mathrm{P}$ as specified by \eqref{eq:CondNormal},
\eqref{eq:values} and
`training' class 0 prevalence $\mathrm{P}[Y=0] = 0.5$.
\item In principle, $(x_{1, te}, y_{1, te}), \ldots,$ $(x_{n, te}, y_{n, te})$ is
an iid sample from $\mathrm{Q}$ as specified by \eqref{eq:CondNormal},
\eqref{eq:values} and
`test' class 0 prevalences $\mathrm{Q}[Y=0] \in$ $\{0.01, 0.05, 0.1, 0.3, 0.5, 0.7, 0.9,$ $0.95, 0.99\}$.
\item However, following the precedent of \citet{Xue:2009:QSC:1557019.1557117}, 
for both datasets we have used stratified sampling such that the proportion of $(x_{i, tr}, 
y_{i, tr})$ with $y_{i, tr} = 0$ in the training set is exactly $\mathrm{P}[Y=0]$, 
and the proportion of $(x_{i, te}, 
y_{i, te})$ with $y_{i, te} = 0$ in the test set is exactly $\mathrm{Q}[Y=0]$.
\end{itemize}
The sample sizes for both the training and the test set samples have been chosen to be 10,000, i.e.
\begin{equation}\label{eq:sizes}
m \ = \ n \ = \ 10,000.
\end{equation}

\begin{table}
\begin{center}
{\small\caption{\textbf{Class 0 prevalence estimates} on the test sets. 
Training set: Binormal with equal variances. Test sets: Binormal with equal variances. `Q[Y=0]' is 
the true test set prevalence of class 0. See Definition~\ref{de:acronyms} for the other acronyms.}\label{tab:3}}
\vspace{1ex}
\begin{tabular}{|l||r|r|r|r|r|r|r|r|r|}\hline
Q[Y=0]&0.01&0.05&0.10&0.30&0.50&0.70&0.90&0.95&0.99 \\ \hline\hline
\multicolumn{10}{|c|}{Prevalence estimates on samples}\\\hline
CDE1&0.1637&0.1910&0.2288&0.3639&0.4989&0.6351&0.7720&0.8014&0.8356 \\ \hline
CDE2&0.0402&0.0682&0.1127&0.2991&0.4986&0.7014&0.8851&0.9230&0.9619 \\ \hline
CDE$\infty$&0.0000&0.0000&0.0040&0.2416&0.4985&0.7653&0.9929&1.0000&1.0000 \\ \hline
ACC&-0.0010&0.0391&0.0947&0.2935&0.4921&0.6924&0.8938&0.9370&0.9873 \\ \hline
EM&0.0070&0.0475&0.0968&0.2988&0.4967&0.6981&0.9007&0.9467&0.9890 \\ \hline\hline
\multicolumn{10}{|c|}{Prevalence estimates on populations}\\\hline
CDE1&0.1655&0.1928&0.2269&0.3635&0.5000&0.6365&0.7731&0.8072&0.8345 \\ \hline
CDE2&0.0406&0.0715&0.1131&0.2994&0.5000&0.7006&0.8869&0.9285&0.9594 \\ \hline
CDE$\infty$&0.0000&0.0000&0.0121&0.2389&0.5000&0.7611&0.9879&1.0000&1.0000 \\ \hline
ACC&0.0100&0.0500&0.1000&0.3000&0.5000&0.7000&0.9000&0.9500&0.9900 \\ \hline
EM&0.0100&0.0500&0.1000&0.3000&0.5000&0.7000&0.9000&0.9500&0.9900 \\ \hline
\end{tabular}
\end{center}
\end{table}

In our experimental design, the sampling is conducted mainly for illustration purposes because at the same time
we also calculate the results at population (i.e.\ sample size $\infty$) level such that we know the
theoretical outcomes. Therefore, for each parametrisation 
of each model, there is no repeated sampling, i.e.\ only one sample is created.

Table~\ref{tab:3} shows the class 0 prevalence estimates made in the double binormal setting of this section. 
Note that in the lower panel of the table, the population estimates by ACC and EM are exact -- as they should be
since in Sections~\ref{se:ACC} and \ref{se:EM} we have proved that both estimators are Fisher consistent 
for the prevalence of class 0 in the family of prior probability shifted distributions. The numbers from the lower panel
also show that, in general, neither of the three CDE-Iterate\footnote{%
Note that in any case the tabulated estimates 
by CDE1, CDE2 and CDE$\infty$ confirm the monotonicity statement in Proposition~\ref{pr:CDE} of Appendix~\ref{se:mixture}
for the convergence of CDE-Iterate.} variants CDE1, CDE2 and CDE$\infty$ are Fisher consistent
for class 0 prevalence under prior probability shift, except for the case of identical training and test set distributions. 

As mentioned in Section~\ref{se:Bayes}, in the setting of the binormal model with equal variances, the Bayes classifier is
unique, irrespective of its specific representation. 
Hence, in this example, there is no chance to work-around the lack of Fisher consistency for CDE-Iterate by
trying to find alternative Bayes classifiers.

However, from the sample estimation numbers in the upper panel of Table~\ref{tab:3},
we can conclude that in practice CDE2 may provide estimates of the class 0 prevalence that are better or at least not
much worse than the ACC and EM estimates.  Table~\ref{tab:4} in Appendix~\ref{se:tables} with the relative errors
of the estimates confirms this observation. Such incidences could explain the favourable performance of CDE-Iterate
observed by \citet{Xue:2009:QSC:1557019.1557117} and \cite{karpov-porshnev-rudakov:2016:SemEval}.

For the sake of completeness, in Tables~\ref{tab:1} and \ref{tab:2} in Appendix~\ref{se:tables}, 
we report the classification accuracies and F-measures respectively, 
for the training set Bayes classifier with adapted thresholds according to \eqref{eq:transform}, 
computed on the different test sets.
The metrics classification accuracy and F-measure here are of interest because \citet{Xue:2009:QSC:1557019.1557117} used
them for measuring the performance of the classifiers discussed in their study. For any classifier 
$g: \mathcal{X} \to \{0,1\}$, we make use of the following
population level formulae for classification accuracy and F-measure:
\begin{subequations}
\begin{gather}\label{eq:accuracy}
\text{Classification accuracy} \ = \ 1 - \text{Classification Error} \ =\ \mathrm{Q}[g(X) = Y], \\
\text{F-measure} \ =\  \frac{2 \times \text{Recall} \times \text{Precision}}{\mathrm{Recall} + \text{Precision}} \ = \
    \frac{2\,\mathrm{Q}[g(X)=0\,|\,Y=0]\,\mathrm{Q}[Y=0\,|\,g(X)=0]}{\mathrm{Q}[g(X)=0\,|\,Y=0]+\mathrm{Q}[Y=0\,|\,g(X)=0]}. \label{eq:Fmeasure}
\end{gather}
\end{subequations}
In these formulae, $\mathrm{Q}$ is used to indicate the test set probability distribution in accordance with the general
assumption of this paper.

Basically, Tables~\ref{tab:1} and \ref{tab:2} in Appendix~\ref{se:tables} demonstrate that it is hard to draw conclusions
on quantification performance by looking at classification accuracy or F-measure as performance metrics. 
Both at sample and at population level, the tabulated classification accuracy values of 
CDE$\infty$, ACC and EM are almost indistinguishable. While the slightly better accuracy values for EM-algorithm and ACC
compared to CDE$\infty$ at population level are indeed sure evidence of better performance, the better values
at sample level might just be random effects.
A similar statement applies to the 
F-measure values of CDE2, ACC and EM. The NaNs in Table~\ref{tab:2}
are caused by zero and negative estimates of the class 0 prevalence.

\begin{figure}[t!p]
{\small\caption{Training set and test set class-conditional densities for Section~\ref{se:binormalInvariant}.
The ratio of the training set densities and the ratio of the test set densities are equal.}\label{fig:densities}}
\begin{center}
\ifpdf
	\resizebox{\height}{9cm}{\includegraphics[width=18cm]{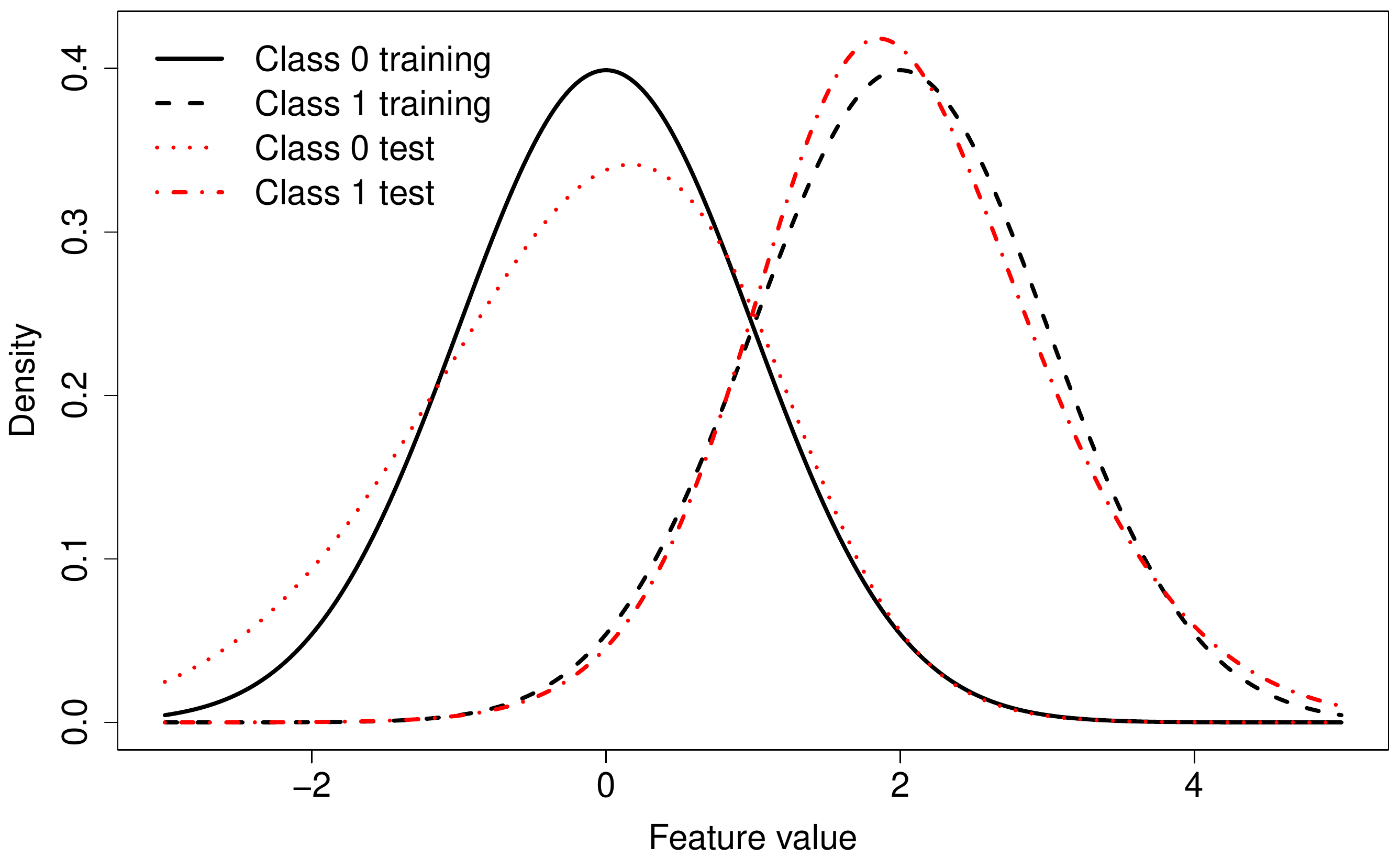}}
\fi
\end{center}
\end{figure}

\subsection{Training set: binormal; test set: non-normal densities, binormal density ratio}
\label{se:binormalInvariant}


The combination of a binormal model with equal variances for the training set 
and a model with the same density ratio but non-normal class-conditional feature distributions
provides an example that fits conveniently into the `invariant density ratio'-type dataset shift setting of Section~\ref{se:quant}. 
We again specify the two models by their class-conditional feature densities.

\textbf{Training set.} Both class-conditional feature distributions are normal, with equal variances, as specified in 
\eqref{eq:CondNormal} and \eqref{eq:values}.

\textbf{Test set.} We specify the test set distribution by class-conditional feature densities $h_0$ and $h_1$ chosen in such
    a way that their ratio $\frac{h_0(x)}{h_1(x)}$, $x \in\mathbb{R}$, is given by \eqref{eq:ratioBinorm}. 
    Then it equals the feature density ratio in the double binormal model from Section~\ref{se:2binormal}. In addition,
    we require that the resulting model is easy to handle numerically but still not too close to the 
    double binormal model. To achieve this, we apply the following steps:
\begin{itemize}
\item We start with a normal density $h^\ast$ characterised by the parameters
\begin{subequations}
    \begin{equation}
	\text{Mean} \ = \ \vartheta, \qquad \text{variance} \ = \ \tau^2\ >\ 0.
	\end{equation}
	For the purpose of this paper, we have chosen
	\begin{equation}\label{eq:values2}
	\vartheta \ = \ 0.5, \qquad \tau \ = \ 1.4.
	\end{equation}
	\end{subequations}
\item Then we apply Theorem~3 of \citet{tasche2014exact} to decompose $h^\ast$ into a mixture $q^\ast\,h_0 + (1-q^\ast)\,h_1$
of $h_0$ and $h_1$, subject to the condition
$\frac{h_0(x)}{h_1(x)} = \exp\left(x\,\tfrac{\mu-\nu}{\sigma^2} + \tfrac{\nu^2-\mu^2}{2\,\sigma^2}\right)$,
$x \in \mathbb{R}$, with $\mu$, $\nu$ and $\sigma$ given by \eqref{eq:values}. The important step for the
decomposition is to determine $q^\ast$. This can be done as suggested in Section~\ref{se:EM}, by solving a version of
Equation~\eqref{eq:EM.result} for the variable $q$:
\begin{subequations}
\begin{equation}\label{eq:par}
0 \ = \ \int_{-\infty}^\infty \frac{R(x)-1}{1+q\,(R(x)-1)} h^\ast(x)\,d x,
\end{equation}
where $R(x)$ is given by \eqref{eq:ratioBinorm}. There is a unique solution $0<q=q^\ast<1$ if and only if
$$\int R(x)\,h^\ast(x)\,dx > 1\quad \text{and} \quad\int R(x)^{-1}\,h^\ast(x)\,dx > 1.$$
With parameters set as in \eqref{eq:values} and \eqref{eq:values2}, the solution for $q$ is
$$q^\ast \ = \ 0.7239184.$$
\item If $q^\ast$ denotes the solution of \eqref{eq:par} the resulting class conditional feature densities 
are determined as
\begin{equation}\label{eq:densities}
h_0(x) \ = \ \frac{R(x)\,h^\ast(x)}{1+q^\ast\,(R(x)-1)} \quad\text{and}\quad 
h_1(x) \ = \ \frac{h^\ast(x)}{1+q^\ast\,(R(x)-1)}, \qquad x \in \mathbb{R}.
\end{equation}
\end{subequations}
\end{itemize}
Once the densities $h_0$ and $h_1$ have been made available by \eqref{eq:densities}, the test set class-conditional 
feature distributions $\mathrm{Q}(X\,|\,Y=0)$ and $\mathrm{Q}(X\,|\,Y=1)$ can be defined as the distributions 
determined by these densities. Figure~\ref{fig:densities} shows the test set class-conditional densities from \eqref{eq:densities}
and the normal densities related to the training set. 

\begin{table}
\begin{center}
{\small\caption{\textbf{Class 0 prevalence estimates} on the test sets. 
Training set: Binormal with equal variances. Test sets: Non-normal densities, binormal density ratio. `Q[Y=0]' is 
the true test set prevalence of class 0. See Definition~\ref{de:acronyms} for the other acronyms.}\label{tab:5}}
\vspace{1ex}
\begin{tabular}{|l||r|r|r|r|r|r|r|r|r|}\hline
Q[Y=0]&0.01&0.05&0.10&0.30&0.50&0.70&0.90&0.95&0.99 \\ \hline\hline
\multicolumn{10}{|c|}{Prevalence estimates on samples}\\\hline
CDE1&0.1548&0.1865&0.2207&0.3526&0.4931&0.6136&0.7508&0.7850&0.8161 \\ \hline
CDE2&0.0313&0.0644&0.1014&0.2773&0.4901&0.6736&0.8728&0.9228&0.9542 \\ \hline
CDE$\infty$&0.0001&0.0010&0.0236&0.2084&0.4848&0.7462&0.9994&1.0000&1.0000 \\ \hline
ACC&-0.0141&0.0325&0.0828&0.2768&0.4835&0.6608&0.8626&0.9129&0.9587 \\ \hline
EM&0.0074&0.0467&0.0957&0.2932&0.4994&0.6853&0.8990&0.9503&0.9913 \\ \hline \hline
\multicolumn{10}{|c|}{Prevalence estimates on populations}\\\hline
CDE1&0.1641&0.1907&0.2240&0.3572&0.4904&0.6236&0.7568&0.7901&0.8167 \\ \hline
CDE2&0.0359&0.0653&0.1049&0.2855&0.4853&0.6890&0.8794&0.9217&0.9531 \\ \hline
CDE$\infty$&0.0000&0.0032&0.0264&0.2168&0.4794&0.7731&1.0000&1.0000&1.0000 \\ \hline
ACC&0.0080&0.0470&0.0958&0.2908&0.4859&0.6810&0.8761&0.9249&0.9639 \\ \hline
EM&0.0100&0.0500&0.1000&0.3000&0.5000&0.7000&0.9000&0.9500&0.9900 \\ \hline
\end{tabular}
\end{center}
\end{table}

For the population-related calculations and the sample simulations, we use the sample sizes as specified in \eqref{eq:sizes},
training set class 0 prevalence $\mathrm{P}[Y=0] = 0.5$ and test set class 0 prevalences  
$\mathrm{Q}[Y=0] \in \{0.01, 0.05, 0.1, 0.3, 0.5, 0.7, 0.9, 0.95, 0.99\}$ as in the double binormal case of 
Section~\ref{se:2binormal}. Again, the samples are stratified with separate sampling from classes 0 and 1.
This is straightforward for the binormal model of the training set, but less straightforward for the test set
distributions given by the class-conditional densities \eqref{eq:densities}. For this sampling 
we have applied the simple accept-reject algorithm suggested by \citet[][Corollary~2.17]{robert2004monte}.

Table~\ref{tab:5} shows the class 0 prevalence estimates made in the 
`binormal -- non-binormal with binormal density ratio' setting of this section. 
In the lower panel of the table, the population estimates by EM are exact -- as they should be
since in Section~\ref{se:EM} we have proved that the EM-algorithm is Fisher consistent 
for the prevalence of class 0 in the family of test set distributions subject to `invariant density ratio'-type
dataset shift. The numbers from the lower panel
also show that, in general, neither ACC nor any of the three CDE-Iterate variants CDE1, CDE2 and CDE$\infty$ are Fisher consistent
for class 0 prevalence under `invariant density ratio'-type dataset shift, not even in the case of training and test sets with
equal class 0 prevalences. However, while the performance of CDE$\infty$ and CDE1 is really poor, CDE2 at least is not
worse than ACC in this setting.

This observation is confirmed by looking at the relative error table \ref{tab:6} in Appendix~\ref{se:tables} and 
the upper `sample' panel of Table \ref{tab:5}. Hence, outside of the prior probability setting, CDE2 may well 
outperform ACC. The `sample' error figures also demonstrate that while in theory the EM-algorithm should 
deliver unbiased estimates of the class 0 test set prevalence under `invariant density ratio'-type dataset shift,
in practice for extreme prevalences like 1\% or 99\% the estimation error may be significant also for the EM-algorithm.

Since we have already seen in Section~\ref{se:2binormal} that measurements of accuracy and F-measure are not very helpful for 
assessing quantification accuracy, we do not present them for the model of this section.

\begin{figure}[t!p]
{\small\caption{Training set and test set class-conditional densities for Section~\ref{se:squareRoot}.
The ratio of the test set densities is equal to the square root of the ratio of the training set densities.}\label{fig:densities5}}
\begin{center}
\ifpdf
	\resizebox{\height}{9cm}{\includegraphics[width=18cm]{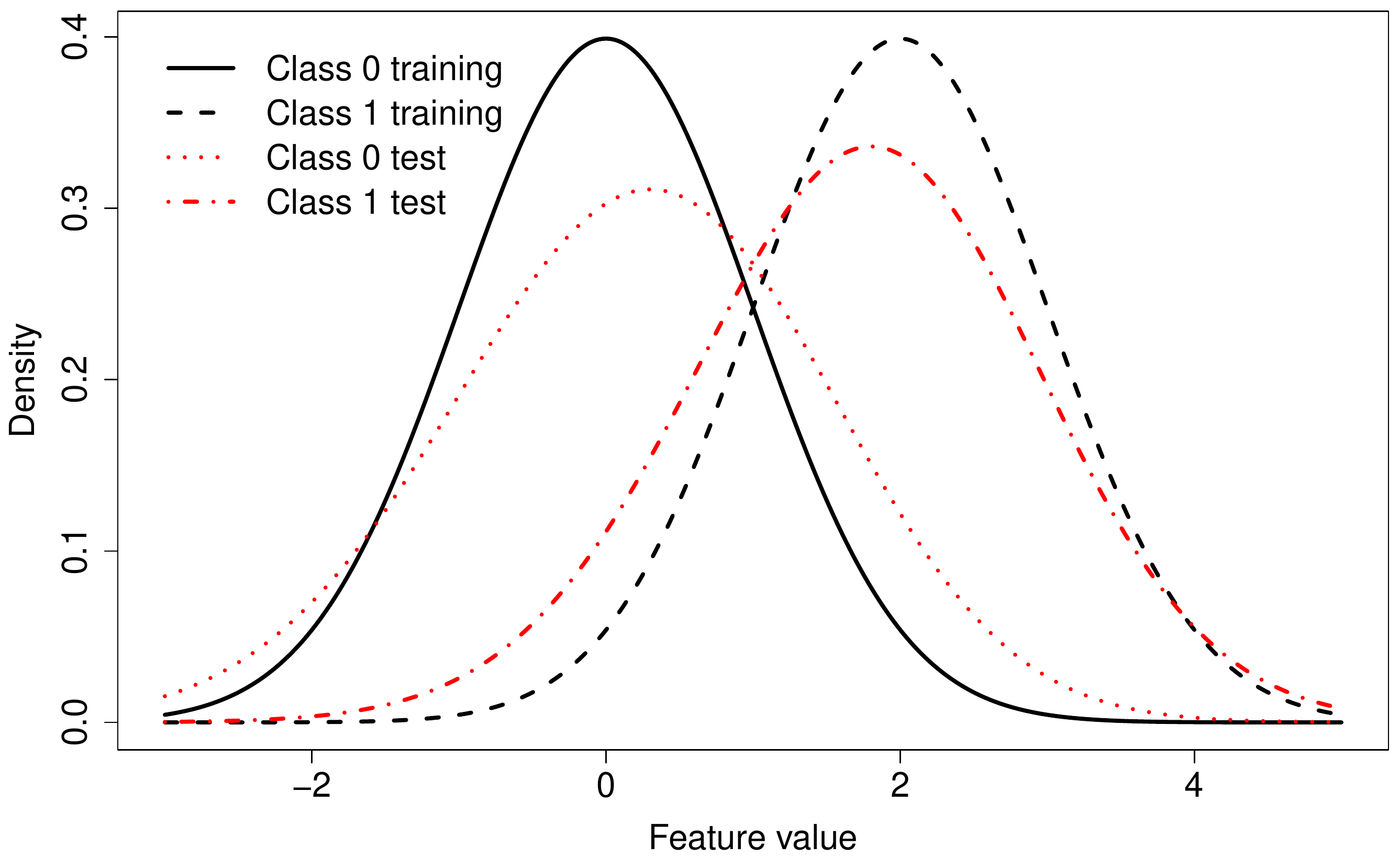}}
\fi
\end{center}
\end{figure}

\subsection{Training set: binormal; test set: non-normal densities, non-binormal density ratio}
\label{se:squareRoot}

The setting of this section for the training and test set distributions is exactly the same as in Section~\ref{se:binormalInvariant},
with the exception that the ratio $\frac{h_0}{h_1}$ of the class-conditional feature densities is not equal
to the test set density as given by \eqref{eq:ratioBinorm} but to its square root:
\begin{equation}\label{eq:squareRoot}
\frac{h_0(x)}{h_1(x)} \ = \ \sqrt{R(X)} \ = \ \exp\left(\frac{2\,x\,(\mu-\nu)+ \nu^2-\mu^2}{4\,\sigma^2}\right).
\end{equation}
Otherwise, the class-conditional feature densities $h_0$ and $h_1$ are again determined by the solution of \eqref{eq:par}  
and \eqref{eq:densities} (with $R(x)$ replaced by $\sqrt{R(X)}$). This time, the solution for $q$ is
$$q^\ast \ = \ 0.8152434.$$
Figure~\ref{fig:densities5} shows 
the test set class-conditional densities from \eqref{eq:densities} and the normal densities related to the training set in this case. 
Also the sample simulations are conducted in the
same way as in Section~\ref{se:binormalInvariant}, again with an accept-reject algorithm deployed for the
test set simulations.

Table~\ref{tab:7} and Table~\ref{tab:8} in Appendix~\ref{se:tables} show that the EM-algorithm is no longer Fisher consistent 
for the prevalence of class 0 when the test set distribution under consideration has not been generated from the training set distribution
by `invariant density ratio'-type dataset shift. Unsurprisingly, given that ACC is not Fisher consistent on test sets which
are not generated by prior probability shift, it is not Fisher consistent in the family of test set distributions generated 
by this modified `invariant density ratio'-type dataset shift either. 
While both CDE1 and CDE2 perform really poorly in this section's model setting,
CDE$\infty$ performs quite well, both at sample and at population level. At sample level, even the CDE$\infty$-estimates 
of the small class 0 prevalences look sensible. 


\section{Conclusions}
\label{se:concl}

In this paper, we have discussed the notion of Fisher consistency as a basic unbiasedness requirement for class prevalence quantifiers
in the presence of dataset shift. The usefulness of Fisher consistency has been demonstrated with three examples of classifiers 
serving as quantifiers: Adjusted Classify \& Count, EM-algorithm, and CDE-Iterate. We have shown by example that CDE-Iterate is not
Fisher consistent even for simple prior probability shift. Adjusted Classify \& Count and EM-algorithm are Fisher consistent
for prior probability shift but lose this property under dataset shifts deviating not much from prior probability shift. 
Hence before relying on prevalence estimates by Adjusted Classify \& Count or EM-algorithm, users should carefully check what
kind of dataset shift they are confronted with. As a further contribution to quantification-related research, we have
suggested a  method, based on the concept of `invariant density ratio'-type
dataset shift, for conveniently generating non-trivial dataset shift beyond
prior probability shift and covariate shift but close to both of these types of dataset shift. 

\begin{table}
\begin{center}
{\small\caption{\textbf{Class 0 prevalence estimates} on the test sets. 
Training set: Binormal with equal variances. Test sets: Non-normal densities, non-binormal density ratio. `Q[Y=0]' is 
the true test set prevalence of class 0. See Definition~\ref{de:acronyms} for the other acronyms.}\label{tab:7}}
\vspace{1ex}
\begin{tabular}{|l||r|r|r|r|r|r|r|r|r|}\hline
Q[Y=0]&0.01&0.05&0.10&0.30&0.50&0.70&0.90&0.95&0.99 \\ \hline\hline
\multicolumn{10}{|c|}{Prevalence estimates on samples}\\\hline
Q[Y=0]&0.01&0.05&0.10&0.30&0.50&0.70&0.90&0.95&0.99 \\ \hline
CDE1&0.2593&0.2766&0.3015&0.3987&0.4926&0.5774&0.6697&0.7007&0.7124 \\ \hline
CDE2&0.1458&0.1666&0.2033&0.3428&0.4883&0.6196&0.7596&0.8083&0.8221 \\ \hline
CDE$\infty$&0.0001&0.0321&0.0920&0.2812&0.4830&0.6612&0.8993&0.9786&0.9998 \\ \hline
ACC&0.1396&0.1650&0.2017&0.3447&0.4828&0.6075&0.7433&0.7889&0.8061 \\ \hline
EM&0.1274&0.1505&0.1952&0.3448&0.4946&0.6251&0.7785&0.8295&0.8482 \\ \hline \hline
\multicolumn{10}{|c|}{Prevalence estimates on populations}\\\hline
CDE1&0.2664&0.2850&0.3081&0.4008&0.4934&0.5861&0.6788&0.7019&0.7205 \\ \hline
CDE2&0.1512&0.1775&0.2109&0.3486&0.4899&0.6320&0.7715&0.8055&0.8324 \\ \hline
CDE$\infty$&0.0000&0.0398&0.0914&0.2886&0.4859&0.6875&0.9020&0.9681&1.0000 \\ \hline
ACC&0.1579&0.1850&0.2189&0.3547&0.4904&0.6261&0.7619&0.7958&0.8230 \\ \hline
EM&0.1307&0.1627&0.2015&0.3500&0.4947&0.6394&0.7875&0.8259&0.8576 \\ \hline
\end{tabular}\end{center}
\end{table}



\addcontentsline{toc}{section}{References}

\newpage
\appendix

\section{Additional tables}
\label{se:tables}

\begin{table}[h]
\begin{center}
{\small\caption{\textbf{Relative error} of class 0 prevalence estimates on the test sets. 
Training set: Binormal with equal variances. Test sets: Binormal with equal variances. `Q[Y=0]' is 
the true test set prevalence of class 0. See Definition~\ref{de:acronyms} for the other acronyms.}\label{tab:4}}
\vspace{1ex}
\begin{tabular}{|l||r|r|r|r|r|r|r|r|r|}\hline
Q[Y=0]&0.01&0.05&0.10&0.30&0.50&0.70&0.90&0.95&0.99 \\ \hline\hline
\multicolumn{10}{|c|}{Relative error of prevalence estimates on samples}\\\hline
CDE1&15.3700&2.8200&1.2880&0.2130&0.0022&0.2163&1.2800&2.9720&15.4400 \\ \hline
CDE2&3.0200&0.3640&0.1270&0.0030&0.0028&0.0047&0.1490&0.5400&2.8100 \\ \hline
CDE$\infty$&1.0000&1.0000&0.9600&0.1947&0.0030&0.2177&0.9290&1.0000&1.0000 \\ \hline
ACC&1.1030&0.2174&0.0527&0.0218&0.0159&0.0253&0.0621&0.2592&0.2651 \\ \hline
EM&0.2987&0.0496&0.0321&0.0040&0.0065&0.0064&0.0073&0.0657&0.1042 \\ \hline\hline
\multicolumn{10}{|c|}{Relative error of prevalence estimates on populations}\\\hline
CDE1&15.5482&2.8558&1.2692&0.2115&0.0000&0.2115&1.2692&2.8558&15.5482 \\ \hline
CDE2&3.0631&0.4304&0.1311&0.0019&0.0000&0.0019&0.1311&0.4304&3.0631 \\ \hline
CDE$\infty$&1.0000&1.0000&0.8789&0.2038&0.0000&0.2038&0.8789&1.0000&1.0000 \\ \hline
ACC&0.0000&0.0000&0.0000&0.0000&0.0000&0.0000&0.0000&0.0000&0.0000 \\ \hline
EM&0.0000&0.0000&0.0000&0.0000&0.0000&0.0000&0.0000&0.0000&0.0000 \\ \hline
\end{tabular}
\end{center}
\end{table}

\begin{table}[h]
\begin{center}
{\small\caption{\textbf{Classification accuracy} on the test sets. 
Training set: Binormal with equal variances. Test sets: Binormal with equal variances. `Q[Y=0]' is 
the true test set prevalence of class 0. See Definition~\ref{de:acronyms} for the other acronyms.}\label{tab:1}}
\vspace{1ex}
\begin{tabular}{|l||r|r|r|r|r|r|r|r|r|}\hline
Q[Y=0]&0.01&0.05&0.10&0.30&0.50&0.70&0.90&0.95&0.99 \\ \hline\hline
\multicolumn{10}{|c|}{Classification accuracy on samples}\\\hline
CDE1&0.9598&0.9414&0.9139&0.8581&0.8430&0.8606&0.9141&0.9334&0.9615 \\ \hline
CDE2&0.9868&0.9591&0.9255&0.8610&0.8431&0.8614&0.9223&0.9581&0.9882 \\ \hline
CDE$\infty$&0.9900&0.9500&0.9040&0.8594&0.8431&0.8573&0.9069&0.9500&0.9900 \\ \hline
ACC&0.9900&0.9591&0.9266&0.8612&0.8427&0.8617&0.9230&0.9593&0.9903 \\ \hline
EM&0.9905&0.9592&0.9265&0.8609&0.8434&0.8613&0.9240&0.9599&0.9903 \\ \hline\hline
\multicolumn{10}{|c|}{Classification accuracy on populations}\\\hline
CDE1&0.9609&0.9397&0.9170&0.8591&0.8413&0.8591&0.9170&0.9397&0.9609 \\ \hline
CDE2&0.9879&0.9588&0.9297&0.8613&0.8413&0.8613&0.9297&0.9588&0.9879 \\ \hline
CDE$\infty$&0.9900&0.9500&0.9109&0.8589&0.8413&0.8589&0.9109&0.9500&0.9900 \\ \hline
ACC&0.9905&0.9595&0.9299&0.8613&0.8413&0.8613&0.9299&0.9595&0.9905 \\ \hline
EM&0.9905&0.9595&0.9299&0.8613&0.8413&0.8613&0.9299&0.9595&0.9905 \\ \hline
\end{tabular}
\end{center}
\end{table}

\begin{table}[h]
\begin{center}
{\small\caption{Classifier \textbf{F-measure} on the test sets. 
Training set: Binormal with equal variances. Test sets: Binormal with equal variances. `Q[Y=0]' is 
the true test set prevalence of class 0. See Definition~\ref{de:acronyms} for the other acronyms.}\label{tab:2}}
\vspace{1ex}
\begin{tabular}{|l||r|r|r|r|r|r|r|r|r|}\hline
Q[Y=0]&0.01&0.05&0.10&0.30&0.50&0.70&0.90&0.95&0.99 \\ \hline\hline
\multicolumn{10}{|c|}{F-measure on samples}\\\hline
CDE1&0.1992&0.5042&0.5952&0.7631&0.8428&0.9005&0.9519&0.9644&0.9803 \\ \hline
CDE2&0.2143&0.4723&0.5509&0.7561&0.8429&0.9033&0.9576&0.9781&0.9940 \\ \hline
CDE$\infty$&NaN&NaN&0.0769&0.7404&0.8429&0.9026&0.9508&0.9744&0.9950 \\ \hline
ACC&NaN&0.4029&0.5360&0.7552&0.8420&0.9032&0.9581&0.9788&0.9951 \\ \hline
EM&0.0952&0.4270&0.5380&0.7559&0.8431&0.9031&0.9587&0.9792&0.9951 \\ \hline\hline
\multicolumn{10}{|c|}{F-measure on populations}\\\hline
CDE1&0.2274&0.5035&0.6106&0.7649&0.8413&0.8994&0.9536&0.9679&0.9799 \\ \hline
CDE2&0.3174&0.4855&0.5815&0.7562&0.8413&0.9030&0.9617&0.9785&0.9939 \\ \hline
CDE$\infty$&NaN&NaN&0.2050&0.7382&0.8413&0.9034&0.9528&0.9744&0.9950 \\ \hline
ACC&0.1697&0.4404&0.5681&0.7563&0.8413&0.9030&0.9619&0.9790&0.9952 \\ \hline
EM&0.1697&0.4404&0.5681&0.7563&0.8413&0.9030&0.9619&0.9790&0.9952 \\ \hline
\end{tabular}
\end{center}
\end{table}

\begin{table}[h]
\begin{center}
{\small\caption{\textbf{Relative error} of class 0 prevalence estimates on the test sets. 
Training set: Binormal with equal variances. Test sets: Non-normal densities, binormal density ratio. `Q[Y=0]' is 
the true test set prevalence of class 0. See Definition~\ref{de:acronyms} for the other acronyms.}\label{tab:6}}
\vspace{1ex}
\begin{tabular}{|l||r|r|r|r|r|r|r|r|r|}\hline
Q[Y=0]&0.01&0.05&0.10&0.30&0.50&0.70&0.90&0.95&0.99 \\ \hline\hline
\multicolumn{10}{|c|}{Relative error of prevalence estimates on samples}\\\hline
CDE1&14.4800&2.7300&1.2070&0.1753&0.0138&0.2880&1.4920&3.3000&17.3900 \\ \hline
CDE2&2.1300&0.2880&0.0140&0.0757&0.0198&0.0880&0.2720&0.5440&3.5800 \\ \hline
CDE$\infty$&0.9900&0.9800&0.7640&0.3053&0.0304&0.1540&0.9940&1.0000&1.0000 \\ \hline
ACC&2.4122&0.3498&0.1718&0.0772&0.0330&0.1307&0.3739&0.7417&3.1336 \\ \hline
EM&0.2638&0.0667&0.0433&0.0227&0.0012&0.0489&0.0096&0.0056&0.1325 \\ \hline \hline
\multicolumn{10}{|c|}{Relative error of prevalence estimates on populations}\\\hline
CDE1&15.4091&2.8146&1.2402&0.1907&0.0192&0.2547&1.4324&3.1988&17.3302 \\ \hline
CDE2&2.5914&0.3051&0.0490&0.0483&0.0295&0.0367&0.2059&0.5655&3.6945 \\ \hline
CDE$\infty$&1.0000&0.9355&0.7365&0.2774&0.0411&0.2437&1.0000&1.0000&1.0000 \\ \hline
ACC&0.2038&0.0604&0.0425&0.0305&0.0281&0.0633&0.2389&0.5024&2.6102 \\ \hline
EM&0.0029&0.0000&0.0000&0.0000&0.0000&0.0001&0.0000&0.0001&0.0010 \\ \hline
\end{tabular}
\end{center}
\end{table}

\clearpage

\begin{table}[t]
\begin{center}
{\small\caption{\textbf{Relative error} of class 0 prevalence estimates on the test sets. 
Training set: Binormal with equal variances. Test sets: Non-normal densities, non-binormal density ratio. `Q[Y=0]' is 
the true test set prevalence of class 0. See Definition~\ref{de:acronyms} for the other acronyms.}\label{tab:8}}
\vspace{1ex}
\begin{tabular}{|l||r|r|r|r|r|r|r|r|r|}\hline
Q[Y=0]&0.01&0.05&0.10&0.30&0.50&0.70&0.90&0.95&0.99 \\ \hline\hline
\multicolumn{10}{|c|}{Relative error of prevalence estimates on samples}\\\hline
CDE1&24.9300&4.5320&2.0150&0.3290&0.0148&0.4087&2.3030&4.9860&27.7600 \\ \hline
CDE2&13.5800&2.3320&1.0330&0.1427&0.0234&0.2680&1.4040&2.8340&16.7900 \\ \hline
CDE$\infty$&0.9900&0.3580&0.0800&0.0627&0.0340&0.1293&0.0070&0.5720&0.9800 \\ \hline
ACC&12.9600&2.3010&1.0168&0.1489&0.0344&0.3082&1.5669&3.2218&18.3881 \\ \hline
EM&11.7408&2.0095&0.9518&0.1495&0.0108&0.2497&1.2146&2.4093&14.1813 \\ \hline \hline
\multicolumn{10}{|c|}{Relative error of prevalence estimates on populations}\\\hline
CDE1&25.6419&4.6990&2.0812&0.3359&0.0131&0.3796&2.2122&4.9611&26.9524 \\ \hline
CDE2&14.1179&2.5494&1.1085&0.1618&0.0201&0.2268&1.2851&2.8896&15.7629 \\ \hline
CDE$\infty$&1.0000&0.2049&0.0863&0.0381&0.0281&0.0418&0.0204&0.3618&1.0000 \\ \hline
ACC&14.7852&2.7000&1.1893&0.1822&0.0192&0.2462&1.3813&3.0839&16.7047 \\ \hline
EM&12.0686&2.2549&1.0149&0.1667&0.0105&0.2019&1.1254&2.4814&13.2365 \\ \hline
\end{tabular}
\end{center}
\end{table}


\section{Proof of the convergence of CDE-Iterate}
\label{se:mixture}

In order to provide a fully rigorous proof of Equation \eqref{eq:limit} that characterises the limit
of CDE-Iterate, we adopt measure-theoretic notation in
this section. See standard textbooks on probability theory like \citet{billingsley3rd} or \citet{Durrett} for reference.

We discuss the problem in a `mixture model' probabilitistic context specified by the following assumption.
\begin{assumption}\label{as:mixModel}
$\mathrm{P}_0$ and $\mathrm{P}_1$ are probability measures on some measurable space $(\Omega, \mathcal{H})$.
Both $\mathrm{P}_0$ and $\mathrm{P}_1$ are absolutely continuous with respect to some measure $\mu$ on $(\Omega, \mathcal{H})$.
The density of $\mathrm{P}_i$ with respect to $\mu$ is $f_i$, $i = 0,1$.
\end{assumption}
Note that in the setting of Sections~\ref{se:shift} and \ref{se:CDE-Iterate}, Assumption~\ref{as:mixModel} is satisfied
when the training set class-conditional feature distributions have got densities. Choose in that case $\Omega = \mathcal{X}$,
$\mathrm{P}_0[H] = \mathrm{P}[X\in H\,|\,Y=0]$ and $\mathrm{P}_0[H] = \mathrm{P}[X\in H\,|\,Y=1]$. The $\sigma$-field 
$\mathcal{H}$ is any appropriate $\sigma$-field on $\mathcal{X}$, for instance the Borel-$\sigma$-field in case $\mathcal{X} = 
\mathbb{R}^d$.

For events $H \in \mathcal{H}$ we denote the complement of $H$ in $\Omega$ by $H^c$, i.e.\ we have $H^c = \Omega \backslash H$.
Then, in the setting of this section, crisp classifiers $g(X)$ with values 0 or 1 are described as events by the
relations $\{g(X) = 0\} = H$ and $\{g(X) = 1\} = H^c$.

The following lemma translates the optimisation problem \eqref{eq:costBayesErr} and its solution \eqref{eq:BayesClassifier}
into this section's notation and enhances them with a statement on the uniqueness of the solution.
\begin{lemma}\label{le:minimum}
Let $a_0, a_1 \ge 0$. Under Assumption \ref{as:mixModel}, then for all $H \in \mathcal{H}$, it holds that
\begin{equation}\label{eq:MinFunc}
a_0\,\mathrm{P}_0[H^c] + a_1\,\mathrm{P}_1[H]\  \ge\ a_0\,\mathrm{P}_0[a_1\,f_1 \ge a_0\,f_0] + 
        a_1\,\mathrm{P}_1[a_1\,f_1 < a_0\,f_0].
\end{equation}
Equality in \eqref{eq:MinFunc} holds
if and only if $0 = \mu\bigl(H\cap \{a_1\,f_1 > a_0\,f_0\}\bigr)$ and $0 = \mu\bigl(H^c\cap \{a_1\,f_1 < a_0\,f_0\}\bigr)$.
\end{lemma} 
\textbf{Proof.} We inspect the following chain of equations and inequalities:
\begin{align*}
a_0\,\mathrm{P}_0[H^c] + a_1\,\mathrm{P}_1[H] & = a_0 + a_1\,\mathrm{P}_1[H] - a_0\,\mathrm{P}_0[H] \\
 & = a_0 + \int_H a_1\,f_1 - a_0\,f_0\, d \mu \\
 & = a_0 + \int_{H\cap \{a_1\,f_1 < a_0\,f_0\}} a_1\,f_1 - a_0\,f_0\, d \mu +
     \int_{H\cap \{a_1\,f_1 > a_0\,f_0\}} a_1\,f_1 - a_0\,f_0\, d \mu\\
 & \stackrel{\text{a)}}{\ge} a_0 + \int_{H\cap \{a_1\,f_1 < a_0\,f_0\}} a_1\,f_1 - a_0\,f_0\, d \mu \\
 & \stackrel{\text{b)}}{\ge} a_0 + \int_{\{a_1\,f_1 < a_0\,f_0\}} a_1\,f_1 - a_0\,f_0\, d \mu \\
 & = a_0\,\mathrm{P}_0[a_1\,f_1 \ge a_0\,f_0] + 
        a_1\,\mathrm{P}_1[a_1\,f_1 < a_0\,f_0]. 
\end{align*}
This proves \eqref{eq:MinFunc}. By inequality a), equality in \eqref{eq:MinFunc} implies 
$0 = \int_{H\cap \{a_1\,f_1 > a_0\,f_0\}} a_1\,f_1 - a_0\,f_0\, d \mu$ and, therefore, 
$0 = \mu\bigl(H\cap \{a_1\,f_1 > a_0\,f_0\}\bigr)$. Similarly, equality in \eqref{eq:MinFunc} implies
$0 = \mu\bigl(H^c\cap \{a_1\,f_1 < a_0\,f_0\}\bigr)$ because of inequality b). \hfill\qed

Lemma~\ref{le:minimum} characterises the solutions $H^\ast$ of the optimisation problem
\begin{equation}\label{eq:probl}
a_0\,\mathrm{P}_0[(H^\ast)^c] + a_1\,\mathrm{P}_1[H^\ast] \ =\ 
\min_{H\in\mathcal{H}}  a_0\,\mathrm{P}_0[H^c] + a_1\,\mathrm{P}_1[H].
\end{equation}
One solution is the event $\{a_1\,f_1 < a_0\,f_0\} \in \mathcal{H}$. However, the solution is not unique.
For instance, $\{a_1\,f_1 \le a_0\,f_0\} \in \mathcal{H}$ is another solution as
it easily can be checked that the two conditions for equality in \eqref{eq:MinFunc} are satisfied. If we
have $\mu(a_1\,f_1 = a_0\,f_0) =0$, then the minimising event $\{a_1\,f_1 < a_0\,f_0\}$ from Lemma~\ref{le:minimum}
is unique in the following sense: If $H^\ast \in \mathcal{H}$ is another minimising event then it follows that
\begin{align*}
0 & \ = \ \mu\bigl(\Delta(H^\ast, \{a_1\,f_1 < a_0\,f_0\})\bigr)\\
& \ = \ \mu\bigl((H^\ast)^c \cap \{a_1\,f_1 < a_0\,f_0\}\bigr) + \mu\bigl(H^\ast \cap \{a_1\,f_1 < a_0\,f_0\}^c\bigr).
\end{align*} 
Hence $H^\ast$ and $\{a_1\,f_1 < a_0\,f_0\}$ are almost everywhere equal.

Assume, similarly to Section~\ref{se:shift}, that there is another probability measure 
$\mathrm{Q}$ on $(\Omega, \mathcal{H})$. $\mathrm{Q}$ is interpreted as the
unconditional distribution of the features on a test set whose class distribution is (not yet) known.
In the notation of this section, then the CDE-Iterate algorithm of \citet{Xue:2009:QSC:1557019.1557117} can be
described as follows:
\begin{itemize}
\item[] \textbf{CDE-Iterate algorithm}
\item[1)] Set initial parameters: $k = 0$, $a_0^{(0)}>0$, $a_1^{(0)}>0$.
\item[2)] Find optimal classifier under Assumption~\ref{as:mixModel}: 
$H_k  =  \{a_1^{(k)}\,f_1 < a_0^{(k)}\,f_0\}$.
\item[3)] Under probability $\mathrm{Q}$ compute
$q_k  =  \mathrm{Q}[H_k]$.
\item[4)] Increment $k$ by 1.
\item[5)] Reset cost parameters: 
$a_0^{(k)} =  q_k$, $a_1^{(k)} =  1-q_k$.
\item[6)] If convergence is reached or $k = k_{\max}$ then stop, else continue with step 2).
\end{itemize}
Convergence of the CDE-algorithm as given above or in the paper by \citet{Xue:2009:QSC:1557019.1557117} is not
obvious. However, we can state the following result.

\begin{proposition}\label{pr:CDE}
Under Assumption~\ref{as:mixModel}, the sequence $(q_k)_{k\ge 0}$ determined by the CDE-algorithm 
as described in this section converges for any probability measure 
$\mathrm{Q}$ on $(\Omega, \mathcal{H})$ and any choice of the intial parameters 
$a_0^{(0)}>0$ and $a_1^{(0)}>0$. The limit $q^\ast = \lim_{k\to\infty} q_k$ satisfies the  equation
\begin{equation}\label{eq:lim}
q^\ast \ =\ 
\begin{cases}
\mathrm{Q}\bigl[(1-q^\ast)\,f_1 \le q^\ast\,f_0,\,
            f_0+f_1 > 0\bigr], & \text{if}\ q_0\ge q_1\
             \text{and}\ q_n > q^\ast\ \text{for all}\ k. \\
\mathrm{Q}\bigl[(1-q^\ast)\,f_1 < q^\ast\,f_0\bigr], & \text{otherwise}.             
         \end{cases}
\end{equation}
\end{proposition}
\textbf{Proof.} Suppose that $q_k \le q_{k+1}$ for some $k$. Then it follows that
\begin{align*}
q_{k+2} & = \mathrm{Q}\bigl[(1-q_{k+1})\,f_1 < q_{k+1}\,f_0\bigr]\\ 
& = \mathrm{Q}\bigl[f_1 < q_{k+1}\,(f_0+f_1)\bigr]\\
& \ge \mathrm{Q}\bigl[(1-q_k)\,f_1 < q_k\,f_0\bigr] \\
& = q_{k+1}.
\end{align*}
Hence $(q_k)_{k\ge 0}$ is non-decreasing if $q_0 \le q_1$. Similarly, it can be shown that 
$(q_k)_{k\ge 0}$ is non-increasing if $q_0 \ge q_1$. It follows that $q^\ast = \lim_{k\to\infty} q_k$ exists.

Note that $\mathrm{Q}\bigl[f_1 < x\,(f_0+f_1)\bigr] =
\mathrm{Q}[f_0+f_1 > 0] \,\mathrm{Q}\bigl[\tfrac{f_1}{f_0+f_1} < x\,|\,f_0+f_1 > 0\bigr]$ is the left-continuous version of a 
distribution function. Therefore, it follows that
\begin{equation*}
\begin{split}
\lim_{y \uparrow x} \mathrm{Q}\bigl[f_1 < y\,(f_0+f_1)\bigr] & = \mathrm{Q}\bigl[f_1 < x\,(f_0+f_1)\bigr], \\
\lim_{y \downarrow x} \mathrm{Q}\bigl[f_1 < y\,(f_0+f_1)\bigr] & = \mathrm{Q}\bigl[f_1 \le x\,(f_0+f_1),\,
            f_0+f_1 > 0\bigr]
\end{split}
\end{equation*}
By definition of $q_k$, this implies \eqref{eq:lim}. \hfill \qed

\end{document}